\documentclass[lettersize,journal]{IEEEtran}
\newcommand{\nn}{AuxSegNet}
\usepackage{graphicx}
\usepackage{amsmath}
\usepackage{amssymb}
\usepackage{hyperref}
\usepackage{adjustbox}
\usepackage{algorithm}
\usepackage{algpseudocode}
\usepackage{multicol}
\usepackage{multirow}
\usepackage{booktabs}
\usepackage{array}
\usepackage{adjustbox}
\usepackage[dvipsnames]{xcolor}
\usepackage{enumitem}

\usepackage{pifont}%
\newcommand{\cmark}{\ding{51}}%
\newcommand{\xmark}{\ding{55}}%

\ifCLASSOPTIONcompsoc
  \usepackage[nocompress]{cite}
\else
  \usepackage{cite}
\fi
\ifCLASSINFOpdf
\else
\fi
\begin{document}
%
\title{Auxiliary Tasks Enhanced Dual-affinity Learning for Weakly Supervised Semantic Segmentation}
%
%
%
%

\author{Lian~Xu,
        Mohammed~Bennamoun,~\IEEEmembership{Senior Member,~IEEE,}
        Farid~Boussaid,
        Wanli~Ouyang,~\IEEEmembership{Senior Member,~IEEE,}
        Ferdous~Sohel,~\IEEEmembership{Senior Member,~IEEE}
        and~Dan~Xu,~\IEEEmembership{Member,~IEEE}
\IEEEcompsocitemizethanks{\IEEEcompsocthanksitem L. Xu and M. Bennamoun and F. Boussaid are The University of Western Australia, Perth, WA, 6009. Emails: \{lian.xu, mohammed.bennamoun, farid.boussad\}@uwa.edu.au.
\IEEEcompsocthanksitem W. Ouyang is with The University of Sydney, SenseTime Computer Vision Group, Australia. Email: wanli.ouyang@sydney.edu.au.
\IEEEcompsocthanksitem F. Sohel is with Murdoch University, Perth, WA. Email: F.Sohel@murdoch.edu.au.
\IEEEcompsocthanksitem D. Xu is with HKUST. Email: danxu@cse.ust.hk.
\IEEEcompsocthanksitem A preliminary version of this work appeared at ICCV 2021~\cite{xu2021leveraging}. Our implementations will be available at~\url{https://github.com/xulianuwa/AuxSegNet+}.
}
}

%
%

\markboth{Journal of \LaTeX\ Class Files,~Vol.~14, No.~8, August~2015}%
{Shell \MakeLowercase{\textit{et al.}}: Bare Demo of IEEEtran.cls for Computer Society Journals}
%



\maketitle

\begin{abstract}
Most existing weakly supervised semantic segmentation (WSSS) methods rely on Class Activation Mapping (CAM) to extract coarse class-specific localization maps using image-level labels. Prior works have commonly used an off-line heuristic thresholding process that combines the CAM maps with off-the-shelf saliency maps produced by a general pre-trained saliency model to produce more accurate pseudo-segmentation labels. We propose AuxSegNet+, a weakly supervised auxiliary learning framework to explore the rich information from these saliency maps and the significant inter-task correlation between saliency detection and semantic segmentation. In the proposed AuxSegNet+, saliency detection and multi-label image classification are used as auxiliary tasks to improve the primary task of semantic segmentation with only image-level ground-truth labels. We also propose a cross-task affinity learning mechanism to learn pixel-level affinities from the saliency and segmentation feature maps. In particular, we propose a cross-task dual-affinity learning module to learn both pairwise and unary affinities, which are used to enhance the task-specific features and predictions by aggregating both query-dependent and query-independent global context for both saliency detection and semantic segmentation. The learned cross-task pairwise affinity can also be used to refine and propagate CAM maps to provide better pseudo labels for both tasks. Iterative improvement of segmentation performance is enabled by cross-task affinity learning and pseudo-label updating. Extensive experiments demonstrate the effectiveness of the proposed approach with new state-of-the-art WSSS results on the challenging PASCAL VOC and MS COCO benchmarks.
\end{abstract}

\begin{IEEEkeywords}
Weakly supervised learning, Semantic segmentation, Auxiliary learning, Affinity learning.
\end{IEEEkeywords}

\vspace{-1em}


%

\section{Introduction}\label{sec:introduction}

%
%
%
%

 

\IEEEPARstart{S}{emantic} segmentation is a fundamental computer vision task, which aims to assign a semantic label to every pixel of an image. It plays an important role in a wide range of applications, such as scene parsing and medical imaging.
Fully supervised semantic segmentation methods have achieved great success with the help of pixel-level dense annotations, which are however expensive and time-consuming to obtain. 
Recently, weakly supervised learning has received considerable attention in various tasks including Re-ID~\cite{wang2020weakly}, action recognition~\cite{zhang2020adapnet}, object localization~\cite{yao2022ts,zhang2022generalized} and detection~\cite{shen2018weakly,zhang2020discriminant,wu2022enhanced}, as it can largely reduce the annotation cost by using weak labels.
The WSSS task commonly uses bounding boxes \cite{hu2018learning,song2019box}, scribbles \cite{lin2016scribblesup,tang2018normalized}, points and image-level labels \cite{pathak2015constrained,kolesnikov2016seed,xu2021atrous,zhang2022componentwise} as weak labels.
This work focuses on using image-level weak supervision, which only provides information on the presence and absence of classes.

\begin{figure}[t]
\setlength{\abovecaptionskip}{0cm}
\setlength{\belowcaptionskip}{0cm}
\centering
{\includegraphics[width=0.43\textwidth]{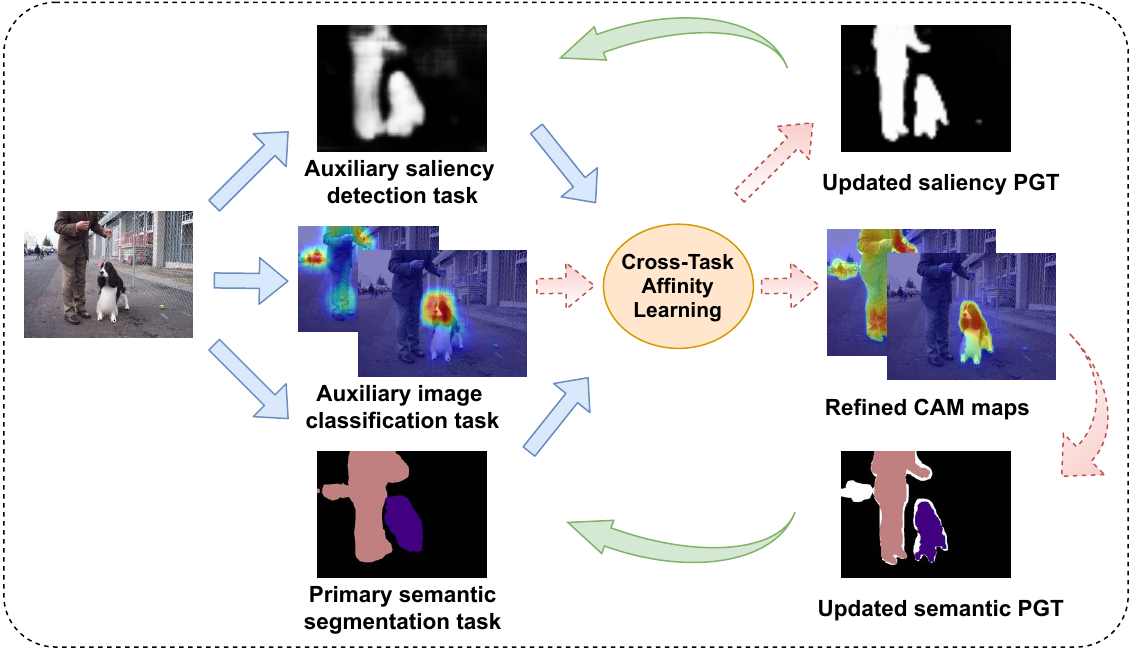}}
  \caption{An illustration of the proposed approach for weakly supervised semantic segmentation. Our approach jointly learns two auxiliary tasks (\textit{i.e.,}~multi-label image classification and saliency detection) and a primary task (\textit{i.e.,}~semantic segmentation) only using image-level ground-truth labels, and performs affinity learning across two dense prediction tasks (\textit{i.e.,}~saliency detection and semantic segmentation). The learned affinity is then used to generate updated pseudo ground-truth (PGT) providing supervision to learn saliency detection and semantic segmentation. } 
\label{teaser}
\vspace{-3ex}
\end{figure}

Most existing WSSS approaches generally include two major steps, \textit{i.e.,} the generation of pseudo segmentation labels and the training of segmentation models. A key tool to generate pseudo segmentation labels is the Class Activation Mapping (CAM)~\cite{zhou2016learning}, which can extract class-specific localization maps (\textit{i.e.,} CAM maps) from CNNs trained on image-level labels. However, the object regions in the raw CAM maps are quite sparse and have very coarse boundaries. To address this problem, various approaches~\cite{wei2017object,wei2018revisiting,jiangintegral,wang2020self} have been developed to improve the CAM maps.
Besides, in order to generate high-quality pseudo segmentation labels, off-the-shelf saliency maps, which are produced by applying a general pre-trained saliency model on the target dataset, have been commonly used in combination with CAM maps to better differentiate object regions from the backgrounds~\cite{chaudhry2017discovering, hou2018self, sun2020mining,zhang2020splitting}. 
However, in previous works, these off-the-shelf saliency maps, which contain coarse but useful object localization information, were only used as fixed binary cues in an off-line pseudo label generation process via heuristic thresholding. They were neither directly involved in the network training nor updated, thereby largely restricting their use and benefit to the segmentation task.

Motivated by the fact that semantic segmentation, saliency detection and image classification are highly correlated, we propose a weakly supervised multi-task deep network (see Fig.~\ref{teaser}), which leverages saliency detection and multi-label image classification as auxiliary tasks to help learn the primary task of semantic segmentation. Through the joint training of these three tasks, an online adaptation can be achieved from pre-trained saliency maps to our target dataset. In addition, the task of saliency detection impels the shared knowledge to emphasize the difference between foreground and background pixels, thus driving the object boundaries of the segmentation outputs to coincide with those of the saliency outputs. Similarly, the image classification task highlights the discriminative features to lead to more accurate segmentation predictions.  

Given that the desired CAM maps should have similar pixel-level object localization information as semantic segmentation and saliency maps, we propose a cross-task affinity learning module which learns the semantic structure from the segmentation and saliency feature maps to guide the propagation of CAM activations. More specifically, two task-specific affinity maps are first learned for the saliency and segmentation tasks, respectively. To capture the complementary information between the two affinity maps, they are then adaptively integrated based on the learned self-attentions to produce a cross-task affinity map. Moreover, as we expect to learn semantic-aware and boundary-aware affinities so as to better update pseudo labels, we impose constraints on learning the cross-task affinities from task-specific supervision and joint multi-objective optimization. The learned cross-task affinity map is further utilized to refine saliency predictions and CAM maps to provide improved pseudo labels for both saliency detection and semantic segmentation respectively, enabling a multi-stage cross-task iterative learning and label updating.

Compared to our earlier work, AuxSegNet~\cite{xu2021leveraging} which proposed a cross-task affinity learning module for modeling the pixel-level pairwise affinity using saliency and segmentation features, the proposed AuxSegNet+ introduces a cross-task dual-affinity learning module which models both pairwise and unary affinities. The pairwise affinity captures the complex correlations between pixel pairs, while the unary affinity encodes the inherent distinctiveness of individual pixels. 
In particular, the proposed framework leverages pairwise affinity to incorporate position-dependent global contexts, enabling pixel-level refinements through context aggregation. Meanwhile, the unary affinity facilitates position-independent global context aggregation, enabling a channel-wise enhancement. 
The extended idea of incorporating the modeling of unary affinity is particularly beneficial for training a weakly supervised segmentation framework for the following critical reasons: (\textbf{i}) The unary affinity provides guided attention based on the global context of the image, such as the boundary of an object in an image. This helps the model focus on important parts of the image. Enhancing certain channels can help direct this attention to regions that are likely to contain the target class, thereby facilitating accurate segmentation. (\textbf{ii}) Limited supervision leads the model to heavily rely on the features it learns. Enhancing specific channels helps the model extract more discriminative and relevant features for the target class and allows the model to learn to emphasize consistent and reliable features despite noisy labels, leading to improved segmentation performance.

In summary, the main contribution is two-fold:
    (\textbf{i}) We propose an effective multi-task auxiliary deep learning framework for weakly supervised semantic segmentation. In the proposed framework, multi-label image classification and saliency detection are used as auxiliary tasks to aid learning of the primary task (i.e., semantic segmentation) using only image-level ground-truth labels. 
    (\textbf{ii}) We propose a cross-task dual affinity learning module, which learns both pixel-level pairwise and unary affinities, to enrich task-specific features, enhance multi-task predictions, and refine pseudo labels for both semantic segmentation and saliency detection. The cross-task dual-affinity learning and the pseudo label updating operate alternately, yielding continuous boosts of the semantic segmentation performance. 

This paper has also made the following additional contributions: (\textbf{i}) We provide a more comprehensive discussion of related works while including a number of additional recent methods. In particular, a systematic review of recent progress in WSSS is presented, focusing on the regularization of weakly supervised segmentation, CAM map generation and CAM map refinement (Sec.~\ref{sec:related_work}). (\textbf{ii}) We elaborate on the construction of the cross-task dual-affinity learning module and provide more in-depth discussions regarding motivations and formulations of the proposed method (Sec.~\ref{sec:3.2}). (\textbf{iii}) We conduct more extensive experiments including comparisons with a number of additional competitive methods to prove the superiority of the proposed method (Sec.~\ref{sec:4.2}), and more ablation studies with quantitative results to demonstrate the effectiveness of the key design of the proposed method (Sec.~\ref{sec:4.3}). (\textbf{iv}) A comprehensive analysis of the advantages and limitations (Sec.~\ref{sec:discussion}) of the proposed method is provided with the help of more qualitative results with common failure cases. Additionally, on the basis of our insightful analysis, we outline future potential works and draw  several important conclusions (Sec.~\ref{sec:conclusion}). (\textbf{v}) The proposed method achieves new state-of-the-art results on the PASCAL VOC and MS COCO datasets.


\section{Related Work}
\label{sec:related_work}
In this section, we review recent works from two closely related perspectives, \textit{i.e.,}~weakly supervised semantic segmentation and auxiliary learning for segmentation.
\vspace{-1em}
\subsection{Weakly supervised semantic segmentation.} 
Most existing WSSS approaches rely on CAM maps to generate pseudo-segmentation labels. However, CAM only discovers the most discriminative object regions, resulting in sparse and incomplete object localization maps, which cannot provide good supervision to train a segmentation network. To mitigate the effect of noisy pseudo-segmentation labels,
Jiang~\textit{et al.}~\cite{jiang2023metaseg} proposed MetaSeg, a meta-learning-based method. This approach effectively generates different weights to suppress noisy pixels and highlight the clean ones. 
Several works have also proposed regularization segmentation losses like SEC loss~\cite{kolesnikov2016seed}, CRF loss~\cite{tang2018regularized} and the contrastive loss~\cite{Keuniversal21}. 

Many prior works have focused on improving the quality of CAM-based pseudo-segmentation labels. For instance, some methods have focused on improving different components in the classification pipeline to generate better CAM maps. We discuss this category of methods from the following three aspects of the classification task: \textit{input}, \textit{feature learning}, and objective \textit{loss} functions. (\textbf{i}) From the perspective of the \textit{input}, given that training a classification network with ordinary images leads to incomplete CAM maps, Wei~\textit{et al.}~\cite{wei2017object} proposed a heuristic strategy, \textit{i.e.,} adversarial erasing (AE), which erases the most discriminative parts discovered by the CAM from the original images and then uses these modified images to re-train the classification network, to force the network to discover more non-discriminative object regions. Similar AE-based strategies have been developed in~\cite{hou2018self, li2018tell, kweon2021unlocking}. Lee~\textit{et al.}~\cite{lee2021anti} proposed to manipulate the images through pixel-level perturbation using the corresponding gradients from the classification output, to increase the classification score. This allows CAM to discover more object regions from the manipulated images. A few other methods~\cite{zhang2020causal, su2021context} have focused on the problem of incorrectly localizing frequently co-occurring objects. Zhang~\textit{et al.}~\cite{zhang2020causal} proposed to cut off the relationship between the context prior and the image by incorporating the image-specific context representation, which is approximated by a combination of the class-specific average segmentation masks, into the input of the classification network.  
Su~\textit{et al.}~\cite{su2021context} proposed a data augmentation method, which first extracts the foreground instance segments using an off-the-self WSSS method and then randomly pastes them onto the original images, to decouple the contextual dependencies between objects.
(\textbf{ii}) From the perspective of \textit{feature learning}, previous methods~\cite{wei2018revisiting, xu2020scale, xu2021atrous, yao2021non, li2021group} have focused on enhancing the contextual information of the classification features, to generate better CAM maps.
Wei~\textit{et al.}~\cite{wei2018revisiting} proposed to use dilated convolutions to enlarge the receptive field of the classification network. This helps the propagation of discriminative object information across more image regions, thus yielding more complete CAM maps. Multi-scale learning has been used to incorporate different-scale local contexts in the classification features to improve CAM maps~\cite{xu2020scale, xu2021atrous}. Moreover, Yao~\textit{et al.}~\cite{yao2021non} employed graph convolutions to enrich the classification features used to generate CAM maps by aggregating global contextual information. WS-FCN~\cite{wang2023coupling} proposed the multi-scale and shallow-deep feature fusion modules to improve the segmentation performance. 
Li~\textit{et al.}~\cite{li2021group} proposed to exploit the inter-image context to refine the classification features by performing message passing across a group of images using a graph neural network.
(\textbf{iii}) From the perspective of \textit{loss} functions with image-level labels, it is believed that the conventional classification objective loss cannot guarantee the localization of complete object regions because it can be achieved by a network only attending to the most discriminative object regions~\cite{zhang2020splitting,chang2020weakly}. To address this limitation, prior works~\cite{wang2020self, zhang2021complementary, zhang2020splitting, chang2020weakly, sun2020mining} have developed various regularization objective \textit{loss} functions to optimize the classification network. Wang~\textit{et al.}~\cite{wang2020self} proposed an equivariant regularization to force the CAM feature maps to be consistent under different affine transformations.  Based on the observation that the sum of the CAM maps extracted from a pair of images with complementary hidden patches (CP) has higher information entropy than the CAM maps of the original image, Zhang~\textit{et al.}~\cite{zhang2021complementary} proposed to increase the information in the original CAM maps by minimizing the distance between the original CAM feature maps and the sum of the CAM maps from a pair of CP images. Zhang~\textit{et al}~\cite{zhang2020splitting} proposed discrepancy and intersection losses to enforce two classification branches to localize different object regions while maintaining common discriminative regions. Moreover, Chang~\textit{et al}~\cite{chang2020weakly} proposed a sub-category classification regularization, which drives the network to localize more complete object regions through a more challenging fine-grained classification objective. Sun~\textit{et al}~\cite{sun2020mining} proposed a co-attention classifier for a pair of images to extract the common and the contrastive semantics, which are optimized by the additional classification objectives determined by the shared and unshared classes, respectively.

\vspace{-0.25em}

Another group of works~\cite{huang2018weakly, wang2018weakly, wang2020weakly, ahn2018learning, fan2020cian} have focused on CAM map refinement methods. 
Huang~\textit{et al.}~\cite{huang2018weakly} proposed a deep seeded region growing strategy to propagate the reliable CAM regions based on a pixel similarity criterion determined by the segmentation prediction. Wang~\textit{et al.}~\cite{wang2018weakly} proposed a region classification network, which is trained on superpixels labeled by the initial CAM maps and iteratively updated by the segmentation prediction, to predict classes for images regions as pseudo labels to train the segmentation network. Wang~\textit{et al.}~\cite{wang2020weakly} improved~\cite{wang2018weakly} by replacing the region classification network with an affinity network, which learns local pixel affinities to refine the pseudo segmentation labels. The affinity network is iteratively optimized using the results from a segmentation network. Ahn~\textit{et al.}~\cite{ahn2018learning} proposed AffinityNet, which learns the local affinity between a pair of adjacent pixels, to refine the pseudo segmentation labels. The affinityNet is trained using the labels obtained from the reliable CAM regions. Fan~\textit{et al.}~\cite{fan2020cian} proposed to learn pixel affinities across images in a segmentation network, to refine the segmentation prediction. The refined segmentation prediction is also used to provide additional online supervision to train the segmentation network while using the CAM maps as pseudo labels. 
DFN~\cite{zhang2022componentwise} proposed a dual-feedback network to iteratively update pseudo labels by using the segmentation outputs and a superpixel-based label refinement mechanism.

\begin{figure*}[t]
\setlength{\abovecaptionskip}{0cm}
\setlength{\belowcaptionskip}{0cm}
\begin{center}
\includegraphics[width=0.85\textwidth]{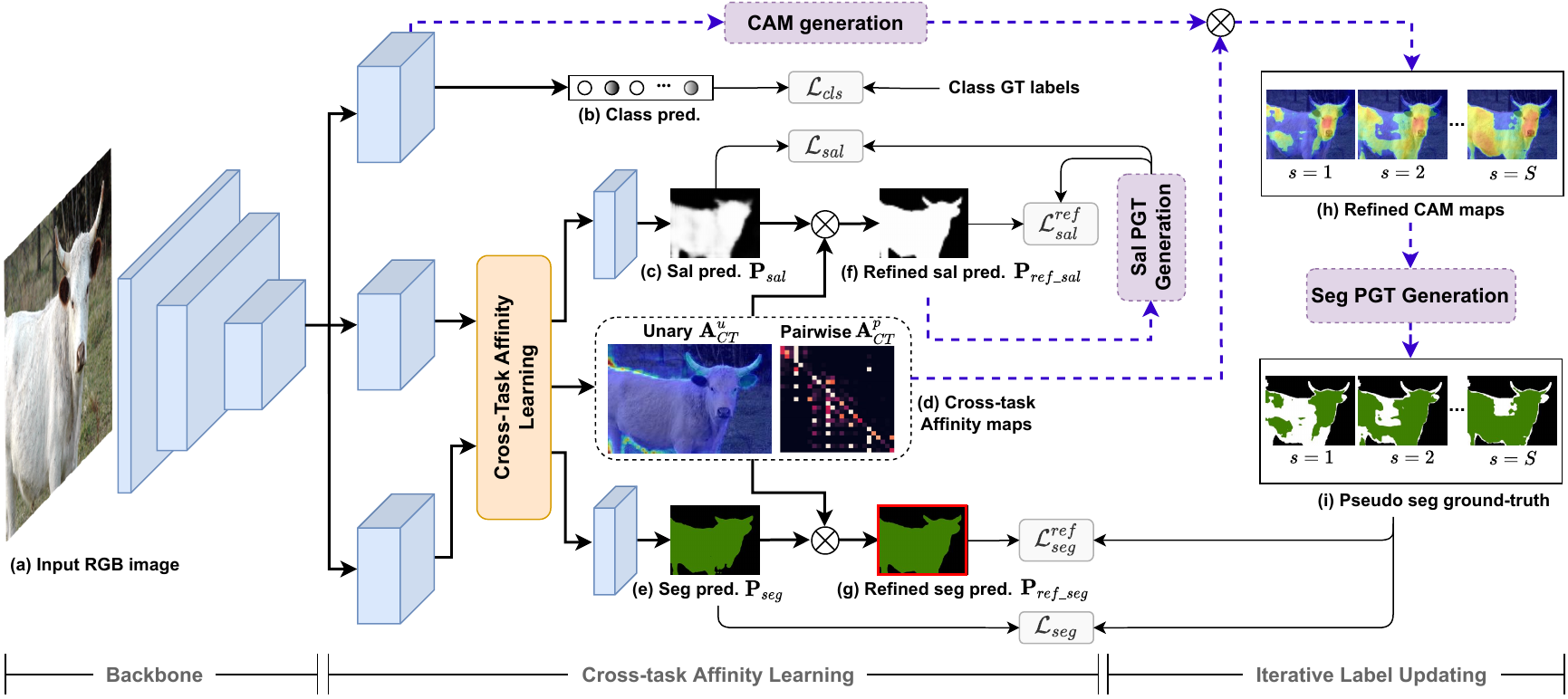}
\end{center}
 \vspace{-0.2cm}
  \caption{An overview of the proposed \nn+. An input RGB image (a) is first passed through a backbone network to extract image features, which are then fed to three branches for multi-label image classification (b), saliency detection (c \& f), and semantic segmentation (e \& g), respectively. The proposed cross-task affinity learning module (see Fig.~\ref{sa_ctal}) takes as inputs the segmentation and saliency feature maps, and outputs enhanced feature maps for predicting both tasks (c \& e) and the cross-task affinity maps (d) including a unary and a pairwise affinity maps for task-specific prediction refinement (f \& g). The refined saliency predictions are used to update pseudo saliency labels, and the learned cross-task pairwise affinity map is used to refine CAM maps (h) to update pseudo segmentation labels (i) to retrain the network. The network training (black solid lines) and label updating (blue dashed lines) are performed alternatively for multiple stages (\textit{i.e.,}~$s=1,2,...,S$) to learn more reliable affinity maps and produce more accurate segmentation predictions. 
  } 
\label{overview}
\vspace{-1.5em}
\end{figure*}

We proposed a weakly supervised multi-task framework, where cross-task pixel-level pairwise affinities are learned to refine CAM maps to generate accurate pseudo-segmentation labels. 
The proposed method is different from those methods that also perform refinement on CAM maps in the following aspects: (\textbf{i}) it learns global pixel affinities across different tasks; (\textbf{ii}) it learns the pixel-level affinities to refine both task-specific representations and predictions; (\textbf{iii}) the affinity can be progressively improved along with more accurate saliency and segmentation results to be achieved with updated pseudo labels on both tasks. Moreover, there are several methods \cite{wei2017object, wang2018weakly, wang2020weakly, zhang2020causal} which also perform iterative refinement on pseudo segmentation labels. In contrast to these methods, which require learning an additional single-modal affinity network to have alternating training with the segmentation network, we perform the cross-task affinity learning simultaneously with the proposed joint multi-task auxiliary learning network.

\vspace{-0.8em}

\subsection{Auxiliary learning for segmentation}
Multi-task learning~\cite{xu2018pad,Sheng2019Unsupervised} allows knowledge sharing across tasks. It has been shown to be more effective in improving the performance of each task, compared to separately training task-specific models. 
In contrast, auxiliary (multi-task) learning aims to improve the performance of the primary task~\cite{XuMoving,liu2019self}. For instance, Dai~\textit{et al.}~\cite{dai2016instance} proposed a multi-task network for segmentation by jointly learning to differentiate instances, estimate masks, and categorize objects. Chen~\textit{et al.}~\cite{chen2016dcan} proposed to improve the segmentation performance by learning an auxiliary contour detection task, with ground-truth labels provided for both primary and auxiliary tasks.

In weakly supervised learning, the joint learning of object detection and segmentation has been explored in~\cite{shen2019cyclic,hwang2021weakly}. The joint learning of image classification and semantic segmentation has been investigated in~\cite{chaudhry2017discovering, zhang2019reliability, araslanov2020single}. Zeng~\textit{et al.}~\cite{zeng2019joint} proposed a multi-task framework to perform fully supervised saliency detection and weakly supervised semantic segmentation.
Lee~\textit{et al.}~\cite{lee2021railroad} proposed to jointly learn image classification and saliency detection by using off-the-shelf saliency maps as saliency labels. These methods only take advantage of jointly learning multiple tasks to obtain better predictions.
In contrast, we leverage two auxiliary tasks (\textit{i.e.,} image classification and saliency detection) to facilitate the feature learning for the primary task of semantic segmentation using image-level labels and off-the-shelf saliency maps. Moreover, we further exploit multi-task features to learn cross-task affinities, which can simultaneously enhance the pseudo labels for both saliency and segmentation tasks to achieve iterative boosts of the segmentation performance.

\section{The Proposed Approach}

\par
\begin{figure*}[t]
\setlength{\abovecaptionskip}{0cm}
\setlength{\belowcaptionskip}{0cm}
\begin{center}
\includegraphics[width=0.8\textwidth]{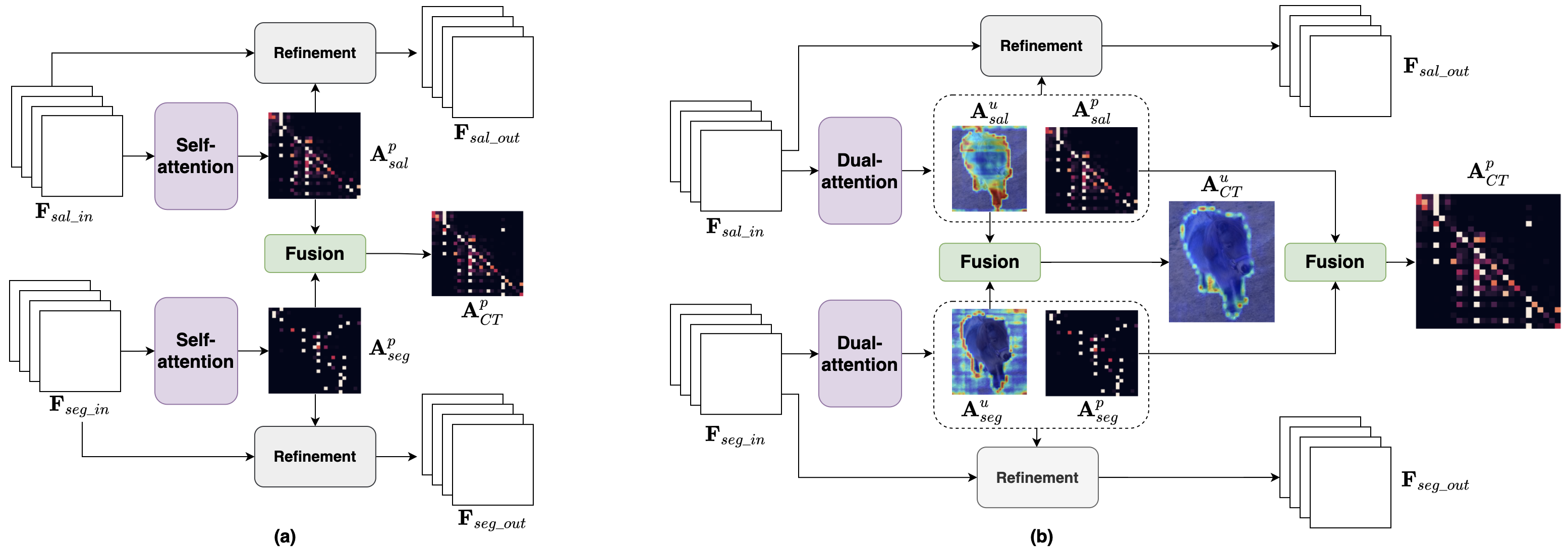}
\end{center}
 \vspace{-0.2cm}
  \caption{The structure of the proposed cross-task affinity learning module. (a) The cross-task pairwise affinity learning module~\cite{xu2021leveraging}, first generates two task-specific pairwise affinity maps by applying the self-attention mechanism on the saliency and segmentation feature maps, respectively. The task-specific pairwise affinity maps are used to refine these corresponding input feature maps. They are further fused to produce a cross-task affinity map. (b) the proposed cross-task dual-affinity learning module, which uses a dual-attention mechanism (detailed in Fig.~\ref{dual_attention}) to generate a unary affinity map and a pairwise affinity map for the saliency and segmentation feature maps, respectively. The task-specific dual affinity maps are used to refine the corresponding feature maps, and these two task-specific dual affinity maps are further fused to produce a cross-task unary affinity map and a cross-task pairwise affinity map, respectively.
  } 
\label{sa_ctal}
\vspace{-1em}
\end{figure*}

\begin{figure}[t]
\begin{center}
\includegraphics[width=0.35\textwidth]{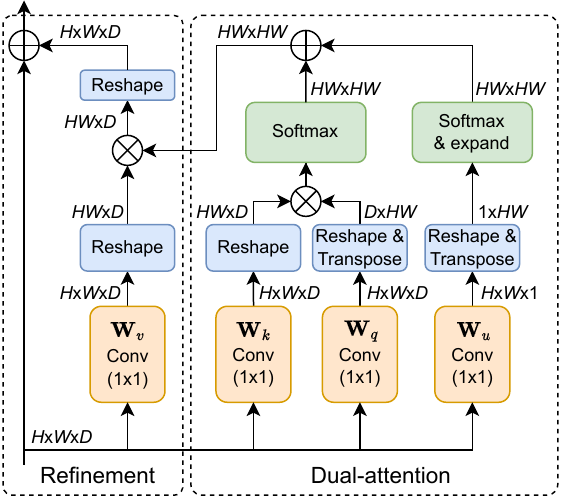}
\end{center}
 \vspace{-0.3cm}
  \caption{Detailed structure of the feature refinement module and the dual-attention mechanism, which captures both the pairwise and unary affinities. $\mathbf{W}_v$, $\mathbf{W}_k$, $\mathbf{W}_q$, and $\mathbf{W}_u$ denote the weight matrices of four $1\times1$ convolutions, respectively.} 
\label{dual_attention}
\vspace{-2em}
\end{figure}

\noindent\textbf{Overview}.~The overall architecture of the proposed \nn+~is shown in Fig.~\ref{overview}. An input RGB image is first fed into a shared backbone network. The generated backbone features are then forwarded to three task-specific branches which predict the class probabilities, a dense saliency map, and a dense semantic segmentation map, respectively.~The proposed cross-task affinity learning module (Fig.~\ref{sa_ctal}) first learns task-specific pixel-level pairwise and unary affinities, which are used to enhance the features of the saliency and segmentation tasks, respectively.
Then, these two task-specific pairwise and unary affinity maps are adaptively integrated via the dual-attention mechanism (Fig.~\ref{dual_attention}), producing cross-task pairwise and unary affinity maps. These affinity maps are further used to refine both the saliency and segmentation predictions during training. The learned cross-task pairwise affinities are used to refine CAM maps, along with the predicted saliency maps, to update the segmentation and saliency pseudo labels, enabling iterative network training and pseudo label updating. Only image-level ground-truth labels and coarse saliency maps are required to train the proposed \nn+.

\vspace{-1em}

\subsection{Multi-Task Auxiliary Learning Framework}
\label{sec:3.1}
\par\noindent\textbf{Auxiliary supervised image classification}. The input image is first passed through the multi-task backbone network to produce a feature map $\mathbf{F} \in \mathbb{R}^{H\times W\times K}$, where $K$ is the number of channels, $H$ and $W$ are the height and width of the map, respectively. A Global Average Pooling (GAP) layer is then applied on $\mathbf{F}$ by aggregating each channel of $\mathbf{F}$ into a feature vector. 
After that, a fully connected (fc) layer is performed as a classifier to produce a probability distribution of the multi-class prediction. Given the weight matrix of the fc classifier $\mathbf{U}\in \mathbb{R}^{K\times C}$, with $C$ denoting the number of classes, the CAM map for a specific class $c$ at a spatial location $(i, j)$ can be formulated as $\mathbf{CAM}^{c}(i, j) = \sum_{k}^{K}\mathbf{U}_{k}^{c}\mathbf{F}_{k}(i,j)$
where $\mathbf{U}_{k}^{c}$ represents the weights corresponding to the class $c$ and the feature channel $k$, and $\mathbf{F}_{k}(i,j)$ represents the activation from the $k$-th channel of $\mathbf{F}$ at a spatial location $(i,j)$. The generated CAM maps are then normalized to be between 0 and 1 for each class $c$ by the maximum value in the two spatial dimensions.

\par\noindent\textbf{Auxiliary weakly supervised saliency detection with label updating.} For the saliency detection branch, feature maps from the backbone network are forwarded to two consecutive convolutional layers with dilated rates of 6 and 12, respectively.~The generated feature maps $\mathbf{F}_{sal\_in}$ are then fed to the proposed cross-task affinity learning module to obtain refined feature maps $\mathbf{F}_{sal\_out}$ and a global cross-task affinity map $\mathbf{A}_{CT}$. $\mathbf{F}_{sal\_in}, \mathbf{F}_{sal\_out} \in \mathbb{R}^{H\times W\times D}$ with $D$ denoting the number of channels. The refined feature maps are used to predict saliency maps $\mathbf{P}_{sal}$ by using a 1$\times$1 convolutional layer followed by a Sigmoid layer. The predicted saliency maps are further refined by the generated cross-task affinity map $\mathbf{A}_{CT}$ to obtain refined saliency predictions $\mathbf{P}_{ref\_sal}$. 
Since no saliency ground truth is provided, we take advantage of pre-trained models which provide coarse saliency maps $\mathbf{Pt}_{sal}$ as initial pseudo labels. For the following stages, we incorporate the refined saliency predictions of the previous stage (\textit{i.e.,}~stage $s-1$) to iteratively perform saliency label updates to continually improve the saliency learning as follows: 
\begin{equation}
      \mathbf{PGT}^{s}_{sal} = \begin{cases}
 \mathbf{\mathbf{Pt}}_{sal} & \text{ if } s = 0, \\ 
 \mathrm{CRF}_d(\frac{\mathbf{P}^{s-1}_{ref\_sal} + \mathbf{Pt}_{sal}}{2})& \text{ if } s > 0,
\end{cases}
\end{equation}
where $\mathbf{PGT}^{s}_{sal}$ denotes the saliency pseudo labels for the $\text{s}^{th}$ training stage, and $\mathrm{CRF}_d(\cdot)$ denotes a densely connected CRF following the formulation in~\cite{hou2019deeply} while using the average of $\mathbf{P}^{s-1}_{ref\_sal}$ and $\mathbf{Pt}_{sal}$ as a unary term.

\par\noindent\textbf{Primary weakly supervised semantic segmentation with label updating.} The semantic segmentation decoding branch shares the same backbone with the saliency decoding branch and the image classification branch. Similar to the saliency decoding branch, two consecutive atrous convolutional layers with rates of 6 and 12 are used to extract task-specific features $\mathbf{F}_{seg\_in}\in \mathbb{R}^{H\times W\times D}$, which are then fed to the cross-task affinity learning module. The output feature maps $\mathbf{F}_{seg\_out}$ are forwarded through a $1\times1$ convolutional layer and a Softmax layer to predict segmentation masks $\mathbf{P}_{seg}$, which are further refined by the learned cross-task affinity map to produce refined segmentation masks $\mathbf{P}_{ref\_seg}$.~To generate pseudo segmentation labels, we follow the conventional procedures \cite{wei2017object,wei2018revisiting,jiangintegral, hou2018self, lee2019ficklenet, sun2020mining} to select reliable object regions from CAM maps and background regions from off-the-shelf saliency maps \cite{hou2019deeply} by hard thresholding. More specifically, to generate the pseudo segmentation labels for the initial training stage (\textit{i.e.,}~stage 0), we first only train the classification branch using image-level labels to obtain the CAM maps. For the following training stages, we generate pseudo-semantic labels by using the CAM maps refined by the cross-task affinities learned at the previous training stage. Moreover, we use the multi-scale strategy to enhance the CAM maps. More specifically, multi-scale inputs are forwarded into the proposed network to generate multiple CAM maps, which are then re-scaled to be of the original size. The multiple CAM maps are then summed up and processed by the min-max normalization to produce the final multi-scale CAM maps.

\vspace{-1em}
\subsection{Cross-task Affinity Learning}
\label{sec:3.2}
We propose a cross-task affinity learning module to enhance both task-specific features and prediction, as well as refine CAM maps.
This work extends the proposed cross-task pairwise affinity learning module (Fig.~\ref{sa_ctal} a) in our earlier work~\cite{xu2021leveraging} to a cross-task dual-affinity learning module (Fig.~\ref{sa_ctal} b) which models both pairwise and unary affinities.
Given an input image, we can obtain two task-specific feature maps from the saliency and segmentation branches of the proposed network, respectively. As these two types of feature maps are both for dense prediction tasks, they are highly correlated in semantic structures and also contain complementary information. It is thus beneficial to leverage the cross-task information to generate more accurate affinities.
Next, we will elaborate on the learning of pairwise and unary affinities and their uses to refine multi-task predictions and CAM maps.
\par\noindent\textbf{Pairwise affinity.} 
To learn the cross-task pairwise affinity, we exploit the self-attention mechanism~\cite{vaswani2017attention} based non-local block to first learn two tasks-specific pairwise affinity maps from the saliency and segmentation features, respectively.
More specifically, given the input saliency feature maps $\mathbf{F}_{sal\_in}, \in \mathbb{R}^{H\times W\times D}$, we use three $1\times1$ convolutional layers to transform the saliency feature maps into the query, key, and value embeddings, which are then spatially flattened to be 2D tensors, \textit{i.e.,} $\mathbf{Q}_{sal}, \mathbf{K}_{sal}, \mathbf{V}_{sal} \in \mathbb{R}^{HW\times D}$. The saliency-specific pairwise affinity matrix $\mathbf{A}^{p}_{sal} \in \mathbb{R}^{HW\times HW}$ is then generated by applying the dot product between each pair of entries of $\mathbf{Q}_{sal}$ and $\mathbf{K}_{sal}$ as: 
\begin{equation}
\begin{aligned}
      \mathbf{A}^{p}_{sal}(i,j) &= \sigma((\mathbf{Q}_{sal}(i)\mathbf{K}_{sal}^T(j))) \\
 &= \frac{\exp{(\mathbf{Q}_{sal}(i)\mathbf{K}_{sal}^T(j))}}{\sum_{n}^{HW}\exp{(\mathbf{Q}_{sal}(i)\mathbf{K}_{sal}^T(n))}},
 \end{aligned}
\end{equation}
where $\sigma$ denotes the softmax function. Each row of $\mathbf{A}^{p}_{sal}$ represents the similarity values of a query position and all spatial positions. Moreover, the affinity matrix $\mathbf{A}^{p}_{sal}$ is reshaped to be a 4D tensor $\mathbf{\hat{A}}^{p}_{sal} \in \mathbb{R}^{H\times W \times H\times W}$, which is further used to enhance the saliency feature maps by adding the refined feature maps $\mathbf{F}^{p}_{sal\_out}$ to the input feature maps as follows:
\begin{equation}
    \mathbf{F}^{p}_{sal\_out}(i,j) = \sum_{k}^{H}\sum_{l}^{W}\mathbf{\hat{A}}^{p}_{sal}(i,j,k,l)\cdot \mathbf{V}_{sal}(k,l), \\
\end{equation}
Similarly, we can obtain the segmentation-specific pairwise affinity map $\mathbf{A}^{p}_{seg}$ and the enhanced segmentation feature maps $\mathbf{F}^{p}_{seg\_out}$ as: 
\vspace{-1em}
\begin{align}
      \mathbf{A}^{p}_{seg}(i,j) &= \sigma((\mathbf{Q}_{seg}(i)\mathbf{K}_{seg}^T(j))),\\
    \mathbf{F}^{p}_{seg\_out}(i,j) &= \sum_{k}^{H}\sum_{l}^{W}\mathbf{\hat{A}}^{p}_{seg}(i,j,k,l)\cdot \mathbf{V}_{seg}(k,l),
\end{align}
where $\mathbf{Q}_{seg}$, $\mathbf{K}_{seg}$ and $\mathbf{V}_{seg}$ are the query, key and value embeddings transformed from $\mathbf{F}_{seg\_in}$ using three $1\times1$ convolutions, respectively. 

To generate a cross-task pairwise affinity map $\mathbf{A}^{p}_{CT}$, the two task-specific affinity maps are then adaptively fused by using a self-attention (SA) module, which consists of two convolutional layers and a softmax layer, and generates the two weight maps, \textit{i.e.,} $\mathbf{W}^{p}_{1}, \mathbf{W}^{p}_{2}\ \in \mathbb{R}^{HW\times HW}$, as follows:
\begin{align}
    [\mathbf{W}^{p}_{1}, \mathbf{W}^{p}_{2}] &= \mathrm{SA}(\mathrm{CONCAT}(\mathbf{A}^{p}_{sal}, \mathbf{A}^{p}_{seg})), \\
    \mathbf{A}^{p}_{CT} &= \mathbf{W}^{p}_{1}*\mathbf{A}^{p}_{sal} + \mathbf{W}^{p}_{2}*\mathbf{A}^{p}_{seg},
\end{align}
where  $\mathrm{CONCAT}$($\cdot$) denotes the concatenation operation.

To investigate what pairwise relationship was learned by the non-local block, one can visualize the corresponding attention map $\mathbf{\hat{A}}(i,j)\in \mathbb{R}^{H\times W}$ given any query position $(i,j)$. Prior works~\cite{cao2019gcnet, yin2020disentangled} observed that the learned attention map of different query positions was similar, indicating that only query-independent global context was learned. This may be due to the lack of guidance for pairwise affinity learning. In contrast, the proposed method uses the pairwise affinity not only to enhance features but also to refine the dense predictions (see Eq.~\ref{ref_sal_p} and Eq.~\ref{ref_seg_p}) for training, during which both the original and refined predictions are supervised by pixel-wise pseudo labels. Each pixel of the refined prediction is a weighted sum of the original prediction, with the weights representing the learned specific affinity values between that query pixel and all pixels. The pixel-wise supervision can directly impact the learning of the pairwise affinity, leading to different global affinity maps for different query positions. The visualization results (Fig.~\ref{affmap}) show that the proposed pairwise affinity captures query position-dependent attention maps.

\par\noindent\textbf{Unary affinity.} We also propose to model unary affinity, which captures position-independent global attention. 
Similar to the cross-task pairwise affinity learning, two task-specific unary affinity maps, \textit{i.e.,} $\mathbf{A}^{u}_{sal}, \mathbf{A}^{u}_{seg} \in\mathbb{R}^{H\times W\times1}$, are first generated from the input saliency and segmentation feature maps, using two $1\times1$ convolutional layers, each of which has 1 output channel, followed by a softmax layer applied in the two spatial dimensions, respectively.
The task-specific unary affinity maps are then reshaped and expanded to be 4D tensors, \textit{i.e.,} $\mathbf{\hat{A}}_{sal}^{u}, \mathbf{\hat{A}}_{seg}^{u}\in\mathbb{R}^{H\times W\times H\times W}$, which are used to aggregate the position-independent global context, \textit{i.e.,} $\mathbf{F}^{u}_{sal\_out},\mathbf{F}^{u}_{seg\_out}$ as:
\vspace{-0.5em}
\begin{align}
    \mathbf{F}^{u}_{sal\_out}(i,j) &=  \sum_{k}^{H}\sum_{l}^{W}\mathbf{\hat{A}}_{sal}^{u}(i,j,k,l)\cdot \mathbf{V}_{sal}(k,l), \\
    \mathbf{F}^{u}_{seg\_out}(i,j) &=  \sum_{k}^{H}\sum_{l}^{W}\mathbf{\hat{A}}_{seg}^{u}(i,j,k,l)\cdot \mathbf{V}_{seg}(k,l).
\end{align}
The aggregated global context can be seen as a vector of channel-wise global responses. Incorporating the unary affinity-guided context enables a channel-wise feature enhancement, similarly to using channel attention~\cite{hu2018squeeze, woo2018cbam, zhang2018occluded, zheng2017learning}. 

The cross-task unary affinity map $\mathbf{A}^{u}_{CT}$ is then obtained by fusing the two task-specific unary affinity maps as:
\begin{align}
    [\mathbf{W}^{u}_{1}, \mathbf{W}^{u}_{2}] &= \mathrm{SA}(\mathrm{CONCAT}(\mathbf{A}^{u}_{sal}, \mathbf{A}^{u}_{seg})), \\
    \mathbf{A}^{u}_{CT} &= \mathbf{W}^{u}_{1}*\mathbf{A}^{u}_{sal} + \mathbf{W}^{u}_{2}*\mathbf{A}^{u}_{seg}.
\end{align}

Finally, the saliency and segmentation feature maps can be enhanced by adding the aggregated position-dependent and position-independent global context captured by the pairwise and unary affinities, to the original input feature maps as:
\begin{align}
    \mathbf{F}_{sal\_out} &= \mathbf{F}^{p}_{sal\_out} + \mathbf{F}^{u}_{sal\_out} + \mathbf{F}_{sal\_in}, \\
    \mathbf{F}_{seg\_out} &= \mathbf{F}^{p}_{seg\_out} + \mathbf{F}^{u}_{seg\_out} + \mathbf{F}_{seg\_in}.
\end{align}

\par\noindent\textbf{Multi-task constraints on cross-task affinity.} To enhance affinity learning, we consider imposing task-specific constraints on the generated cross-task affinity maps. To this end, both the generated unary affinity and pairwise affinity matrices~\textit{i.e.,} $\mathbf{A}^{u}_{CT}$ and $\mathbf{A}^{p}_{CT}$, are 
used to refine both the original saliency and segmentation predictions~\textit{i.e.,} $\mathbf{P}_{sal}$ and $\mathbf{P}_{seg}$, respectively, during training as:
\begin{equation}
\label{ref_sal_u}
    \mathbf{P}^{u}_{ref\_sal}(i,j) = \mathbf{P}_{sal}(k,l) + \sum_{k}^{H}\sum_{l}^{W}\mathbf{A}^{u}_{CT}(i,j,k,l)\cdot \mathbf{P}_{sal}(k,l),
\end{equation}
\begin{equation}
\label{ref_sal_p}
    \mathbf{P}^{p}_{ref\_sal}(i,j) = \sum_{k}^{H}\sum_{l}^{W}\mathbf{A}^{p}_{CT}(i,j,k,l)\cdot \mathbf{P}_{sal}(k,l),
\end{equation}
\begin{equation}
\label{ref_seg_u}
    \mathbf{P}^{u}_{ref\_seg}(i,j) = \mathbf{P}_{seg}(k,l) + \sum_{k}^{H}\sum_{l}^{W}\mathbf{A}^{u}_{CT}(i,j,k,l)\cdot \mathbf{P}_{seg}(k,l),
\end{equation}
\begin{equation}
\label{ref_seg_p}
    \mathbf{P}^{p}_{ref\_seg}(i,j) = \sum_{k}^{H}\sum_{l}^{W}\mathbf{A}^{p}_{CT}(i,j,k,l)\cdot \mathbf{P}_{seg}(k,l).
\end{equation}
The learning of the cross-task affinity can thus gain effective supervision from both saliency and segmentation pseudo labels.~Therefore, the improvements on the updated pseudo labels can boost affinity learning. 

\vspace{1em}
\par\noindent\textbf{CAM map refinement.} The trained pairwise cross-task affinity can be extracted from the proposed framework to refine the CAM maps extracted from the classification branch as:
\begin{equation}
\label{ref_cam}
    \mathbf{CAM}^{c}_{ref}(i,j) = \sum_{k}^{H}\sum_{l}^{W}\mathbf{A}^{p}_{CT}(i,j,k,l)\cdot \mathbf{CAM}^{c}(k,l).
\end{equation}
The refined CAM maps are then used to generate pseudo labels (see Section~\ref{sec:3.1}) for the next training stage.

\noindent\textbf{Time complexity.} 
The standard non-local block that AuxSegNet~\cite{xu2021leveraging} used to model pairwise affinity has a time complexity of $\mathcal{O}(HWD^2 + (HW)^2D)$. In contrast, the proposed dual affinity module which models both pairwise and unary affinities has a time complexity of $\mathcal{O}(HWD^2 + (HW)^2D + HWD + (HW)^2)$, where $H, W, D$ denote the height, the width, and the number of channels of the input feature maps, respectively. The additional computational cost of the proposed dual-affinity module compared to the non-local block is minimal, specifically only 0.8\% for $D$ = 128 in our experiments.

\begin{algorithm}
\textbf{Input:} Images $\mathbf{I}$, image-level class labels $\mathbf{y}$, off-the-shelf saliency maps $\mathbf{Pt}_{sal}$ \\
\textbf{Output:} Segmentation prediction $\textbf{P}_{seg}$ 
\begin{algorithmic}[1]
    \State $M_{cls}.train(\mathbf{I}, \mathbf{y})$  \algorithmiccomment{Train a classification model}
    \State $\mathbf{MSCAM} = M_{cls}.get\_multiscale\_CAM(\mathbf{I})$ 
    \State $\mathbf{PGT}^1_{seg} = combine(\mathbf{MSCAM}, \mathbf{Pt}_{sal})$ \algorithmiccomment{Generate initial segmentation pseudo ground-truth}
    \State $\mathbf{PGT}^1_{sal} = \mathbf{Pt}_{sal}$ \algorithmiccomment{Initial saliency pseudo ground-truth}
    \For {each training stage $i$}
        \State $M_{mt}^i.train(\mathbf{I}, \mathbf{PGT}^i_{seg}, \mathbf{PGT}^i_{sal}, \mathbf{y})$  \algorithmiccomment{Train the proposed weakly supervised multi-task network}
        \If{$i < num\_stages$} 

        \For {each image scale $s$}  \algorithmiccomment{To generate multi-scale refined CAM maps}
        \State $\mathbf{CAM}_s^i = M_{mt}^i.get\_CAM(\mathbf{I}_{s})$ 
        \State $\mathbf{A}^{p,i}_{ct,s} =M_{mt}^i.get\_cross\_task\_affinity(\mathbf{I}_{s})$  
        \State$\mathbf{RCAM}_{s}^i = refine\_CAM(\mathbf{CAM}_s^i, \mathbf{A}^{p,i}_{ct,s})$ 
        \EndFor
        \State $\mathbf{MSRCAM}_i = Norm(\sum_s(\mathbf{RCAM}_s^i))$
        \State$\mathbf{PGT}^{i+1}_{sal} = CRF(\mathbf{P}^i_{sal}, \mathbf{Pt}_{sal})$  \algorithmiccomment{Saliency pseudo ground-truth updating}
        \State$\mathbf{PGT}^{i+1}_{seg} = combine(\mathbf{MSRCAM}^i, \mathbf{PGT}^{i+1}_{sal})$  \algorithmiccomment{Segmentation pseudo ground-truth updating}
        \Else
        \State$\mathbf{P}_{seg} = M^i_{mt}.seg\_predict(\textbf{I})$ \algorithmiccomment{Final segmentation prediction}
        \EndIf
    \EndFor
\caption{Pseudo-code for the proposed weakly supervised semantic segmentation algorithm.}
\label{algorithm}
\end{algorithmic}
\end{algorithm}

\subsection{Training and Inference}
The procedures for training~\nn+, generating pseudo labels and inference are shown in Algorithm~\ref{algorithm}.

\par\noindent\textbf{Overall optimization objective.} The overall objective loss function is the sum of the losses for the three tasks:
\begin{equation}
\begin{gathered}
    \mathcal{L}_{total} = \lambda_1\cdot\mathcal{L}_{cls} +   
    \lambda_2\cdot(\mathcal{L}_{sal}+\mathcal{L}^{ref}_{sal}) + 
    \lambda_3\cdot(\mathcal{L}_{seg}+\mathcal{L}^{ref}_{seg}), \\
    \mathcal{L}^{ref}_{sal}=
    \mathcal{L}^{u}_{sal} + \mathcal{L}^{p}_{sal}, \\
     \mathcal{L}^{ref}_{seg}=
    \mathcal{L}^{u}_{seg} + \mathcal{L}^{p}_{seg}, 
\end{gathered}
\end{equation}
where $\mathcal{L}_{cls}$ is a multi-label soft margin loss computed between the predicted class probabilities and the image-level ground-truth labels to optimize the image classification network branch;
$\mathcal{L}_{sal}$,  $\mathcal{L}^{u}_{sal}$ and $\mathcal{L}^{p}_{sal}$ are the binary cross entropy losses computed between the pseudo saliency label maps and $\mathbf{P}_{sal}$, $\mathbf{P}^{u}_{ref\_sal}$, $\mathbf{P}^{p}_{ref\_sal}$, respectively, to optimize the saliency detection network branch and the affinity fusion module; $\mathcal{L}_{seg}$,  $\mathcal{L}^{u}_{seg}$ and $\mathcal{L}^{p}_{seg}$ are pixel-wise cross entropy losses calculated between the pseudo segmentation label maps and $\mathbf{P}_{seg}$, $\mathbf{P}^{u}_{ref\_seg}$, $\mathbf{P}^{p}_{ref\_seg}$, respectively, and these losses optimize the segmentation branch and the affinity fusion module; the loss weights, \textit{i.e.,}~$\lambda1$, $\lambda2$ and $\lambda3$ are all set to 1 for each task.

\par\noindent\textbf{Stage-wise training.} We use a stage-wise training strategy for the entire multi-task network optimization. First, we only train the image classification branch with image-level labels for 15 epochs. The learned network parameters are then used as initialization to train the entire proposed \nn+. 
We continually train the network for multiple stages. For each training stage, we set the maximum number of epochs to 15 and used the early stopping with the patience of 5 epochs. We observed the network training converged/stopped at around 10 epochs.
The saliency and segmentation pseudo labels are updated after each training stage.

\par\noindent\textbf{Inference.} For inference, we use the segmentation prediction refined by the learned cross-task pairwise affinity map as the final segmentation results.

\section{Experiments}

\subsection{Experimental Settings}
\par\noindent\textbf{Datasets.} To evaluate the proposed method, we conducted experiments on PASCAL VOC 2012 \cite{everingham2010pascal} and MS COCO datasets \cite{lin2014microsoft}. \textbf{PASCAL VOC} has 21 classes (including a background class) for semantic segmentation. This dataset has three subsets, \textit{i.e.,} training (\textit{train}), validation (\textit{val}), and test with 1,464, 1,449, and 1,456 images, respectively. Following common practice, \textit{e.g.}, \cite{chen2014semantic,kolesnikov2016seed}, we used additional data from \cite{hariharan2011semantic} to construct an augmented dataset with 10,582 images for training. \textbf{MS COCO} contains 81 classes (including a background class). It has 80K training images and 40K validation images. Note that only image-level ground-truth labels from these benchmarks are used in the training process. 

\begin{table}[t]
\caption{Evaluation (\%) of the generated initial pseudo segmentation labels using different CAM thresholds on PASCAL VOC.}
\label{CAM_threshold}
\vspace{-1em}
\centering
\small
\resizebox{0.7\linewidth}{!}{
\begin{tabular}{cccc}
\toprule 
CAM threshold&Precision&Recall&mIoU\\
\midrule
0.2& 85.6&\textbf{42.9}&\textbf{40.6}\\
0.25&86.0&38.5&36.7\\
0.3 &86.4&34.4&32.9 \\
0.35& \textbf{86.7} & 30.6&29.4 \\

\bottomrule
\end{tabular}}
\vspace{-1em}
\end{table}

\begin{table}[t]
\caption{Performance comparison of WSSS methods in terms of mIoU(\%) on the PASCAL VOC \textit{val} and \textit{test} sets. Sup.: supervision. B: Bounding-box-level GT. I: image-level GT. S: saliency maps.
\label{sota_res38}}
\vspace{-1em}
\centering
\small
\resizebox{1.0\linewidth}{!}{
\begin{tabular}{lcccc}
\toprule
Method            & Backbone  & Sup.          & Val            & Test           \\ \midrule
MetaSeg (TNNLS23)~\cite{jiang2023metaseg} & VGG19 & B & 68.7 & 69.2 \\
\midrule
AffinityNet (CVPR18) \cite{ahn2018learning}&ResNet38&I&61.7&63.7 \\
  Zhang \textit{et al.}~ (AAAI20) \cite{zhang2019reliability} & ResNet38 & I&62.6 & 62.9 \\
    Luo \textit{et al.}~ (AAAI20)  \cite{luo2020learning}  & ResNet101&I &   64.5  & 64.6 \\
Chang \textit{et al.}~ (CVPR20) \cite{chang2020weakly}  &            ResNet101    &I     &      66.1        &    65.9\\ 
Araslanov \textit{et al.}~ (CVPR20) \cite{araslanov2020single}    &          ResNet38    &I     &      62.7        &   64.3       \\ 
SEAM (CVPR20) \cite{wang2020self} & ResNet38 &I& 64.5 & 65.7 \\
 CONTA (NeurIPS20) \cite{zhang2020causal}   &          ResNet38         &  I&    66.1       &    66.7            \\  
 AdvCAM (CVPR21) \cite{lee2021anti} & ResNet101 & I &68.1 &68.0 \\
 ECS-Net (ICCV21) \cite{sun2021ecs} & ResNet38 & I &66.6 & 67.6 \\
 Kweon \textit{et al.} (ICCV21) \cite{kweon2021unlocking} & ResNet38 & I &68.4&68.2 \\
 CDA (ICCV21) \cite{su2021context} & ResNet38 & I &66.1&66.8 \\
 Zhang \textit{et al.} (ICCV21) \cite{zhang2021complementary} & ResNet38 & I & 67.8&68.5 \\
 PMM~(ICCV21) \cite{li2021pseudo}&ResNet38&I&68.5&69.0 \\
 AMR (AAAI22)~\cite{qin2021activation}&ResNet101&I&68.8&69.1 \\
WS-FCN (TNNLS23)~\cite{wang2023coupling}&ResNet38&I&63.2&64.2 \\
 \midrule
DSRG (CVPR18) \cite{huang2018weakly} &ResNet101 &I+S&61.4&63.2 \\
MCOF (CVPR18) \cite{wang2018weakly} & ResNet101 & I+S &60.3 & 61.2 \\
SeeNet (NeurIPS18) \cite{hou2018self} &ResNet101&I+S&63.1 &62.8 \\
FickleNet (CVPR19)  \cite{lee2019ficklenet}  &ResNet101 &  I+S&64.9 & 65.3 \\
OAA$^{+}$ (ICCV19)  \cite{jiangintegral} & ResNet101 & I+S&   65.2& 66.4  \\
Zeng \textit{et al.}~ (ICCV19) \cite{zeng2019joint}& DenseNet&I+S &63.3&64.3\\ 
CIAN (AAAI20)  \cite{fan2020cian}    &ResNet101& I+S&  64.3 &  65.3 \\
ICD (CVPR20) \cite{fan2020learning}    &          ResNet101          &   I+S&    67.8         &    68.0           \\ 
 Zhang \textit{et al.}~ (ECCV20) \cite{zhang2020splitting}   &       ResNet50  &   I+S&   66.6         &   66.7    \\
 Sun \textit{et al.}~ (ECCV20) \cite{sun2020mining}   &          ResNet101        &I+S&        66.2          &      66.9        \\
 EDAM (CVPR21) \cite{wu2021embedded} & ResNet101 & I+S & \textbf{70.9} & 70.6 \\
 EPS (CVPR21) \cite{lee2021railroad} & ResNet38 & I+S & 68.9&70.0\\
 Yao~\textit{et al.} (CVPR21) \cite{yao2021non} & ResNet101 & I+S &68.3&68.5 \\
 DFN (TNNLS22)~\cite{zhang2022componentwise}&ResNet101 & I+S & 62.6 & 64.0 \\
 
 \midrule
  \textbf{\nn} (ICCV21)~\cite{xu2021leveraging} & ResNet38 &I+S&    69.0 &  68.6\\
   \textbf{\nn+} (Ours) & ResNet38 &I+S&  70.7 &  \textbf{70.9}\\
  \bottomrule
\end{tabular}
}
\vspace{-1em}
\end{table}

\par\noindent\textbf{Evaluation metrics.}
As in previous works \cite{jiangintegral,lee2019ficklenet,wei2018revisiting,huang2018weakly,ahn2018learning,wang2018weakly}, we used the mean Intersection-over-Union (mIoU) of all classes between the segmentation outputs and the ground-truth images to evaluate the performance of our method on the \textit{val} and \textit{test} sets of PASCAL VOC and the \textit{val} set of MS COCO. We also used precision, recall, and mIoU to evaluate the quality of the pseudo-segmentation labels. The results on the PASCAL VOC \textit{test} set were obtained from the official PASCAL VOC online evaluation server.
\par\noindent\textbf{Implementation details.}
We use ResNet38 \cite{wu2019wider,ahn2018learning} as the backbone network. 
For data augmentation, we used random scaling with a factor of $\pm 0.3$, random horizontal flipping, and random cropping to size $321\times 321$. The polynomial learning rate decay was chosen with an initial learning rate of 0.001 and a power of 0.9. We used the stochastic gradient descent (SGD) optimizer to train \nn+~with a batch size of 4.
We followed the previous works~\cite{zhou2016learning, kolesnikov2016seed, huang2018weakly, wei2017object} 
to determine the foreground regions of the normalized CAM maps and the normalized off-the-shelf saliency maps by using the thresholds 0.2 and 0.06, respectively. We used the same thresholds for all datasets. 
The criterion for threshold selection is to achieve a balance between precision and recall. Several works have used various thresholds for CAM maps, including 0.25~\cite{wu2021embedded}, 0.3~\cite{fan2020cian}, and 0.35~\cite{lee2019ficklenet}. However, as shown in Table~\ref{CAM_threshold}, as the CAM threshold increases, the precision is improved slightly while the recall drops significantly, resulting in a poorer mIoU. This suggests that a threshold of 0.2 is the optimal choice for CAM maps. 
At inference, we use multi-scale testing and CRFs with hyper-parameters suggested in \cite{chen2014semantic} for post-processing.


\begin{figure}[t]
\begin{center}
\includegraphics[width=0.4\textwidth]{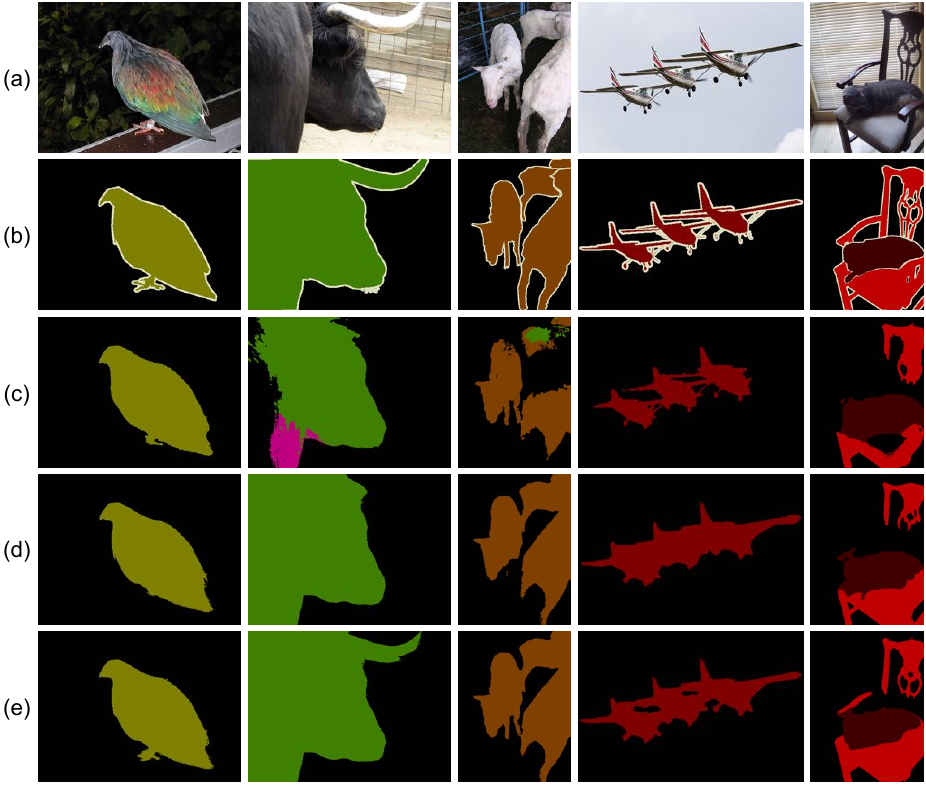}
\end{center}
 \vspace{-0.2cm}
  \caption{Qualitative segmentation results on the PASCAL VOC \textit{val} set. (a) Input images. (b) Ground-truth segmentation masks. (c) Predicted segmentation masks by EPS~\cite{lee2021railroad}. (d) Predicted segmentation masks by~\nn. (e) Predicted segmentation masks by~\nn+.} 
\label{ext-segresults-voc}
\vspace{-1em}
\end{figure}

\subsection{Comparison with State-of-the-arts}
\label{sec:4.2}
\par\noindent\textbf{PASCAL VOC.} We compared the segmentation performance of the proposed method with state-of-the-art WSSS approaches. Table \ref{sota_res38} shows that the proposed \nn+ achieves mIoUs of 70.7\% and 70.9\%, attaining significant improvements of 1.7\% and 2.3\%, compared to \nn, on the \textit{val} and \textit{test} sets, respectively. 
More specifically, although our method uses only image-level ground-truth labels, it outperforms MetaSeg~\cite{jiang2023metaseg}, which uses bounding-box-level ground-truth labels, by margins of 2.0\% and 1.7\% on the \textit{val} and \textit{test} sets, respectively.
Our method outperforms the recent methods \cite{wu2021embedded, lee2021railroad, yao2021non} using off-the-shelf saliency maps by 2.4\%, 0.9\% and 0.3\%, respectively, on the \textit{test} set. 
Compared to DFN~\cite{zhang2022componentwise}, which also performs iterative pseudo-label updating, our proposed method improves mIoUs by significant margins of 8.1\% and 6.9\% on the val and test sets, respectively. 
This can be attributed to the superior effectiveness of our affinity, which is learned from multi-task deep features for refining the object boundaries within pseudo labels. In contrast, DFN~\cite{zhang2022componentwise} learns the affinity from the low-level segmentation features.
The qualitative segmentation results on PASCAL VOC \textit{val} set are shown in Fig.~\ref{ext-segresults-voc}. Our segmentation results are shown to adapt well to different object scales and boundaries in various challenging scenes.  
We can also observe that compared to the two recent methods~\nn~\cite{xu2021leveraging} (Fig.~\ref{ext-segresults-voc}d) and EPS~\cite{lee2021railroad} (Fig.~\ref{ext-segresults-voc}c), the proposed \nn+ (Fig.~\ref{ext-segresults-voc}d) can recover more object details, such as the chair in the last column, and produce more accurate object segmentation even when the object parts are not noticeable, such as the bird feet and the cow horns, which are similar to the background in colors, in the first and second columns.

\begin{table}[t]
\caption{Performance comparison of WSSS methods in terms of mIoU(\%) on the MS COCO \textit{val} set. }
\label{coco}
\vspace{-1em}
\small
\centering
\begin{tabular}{lccc}
\toprule 
Method            & Backbone  &Sup.          & Val                       \\ \midrule
Wang \textit{et al.}~ (IJCV20) \cite{wang2020weakly}&VGG16&I&27.7\\
    Luo \textit{et al.}~ (AAAI20)  \cite{luo2020learning}  &     VGG16              &  I&    29.9 \\
SEAM (CVPR20) \cite{wang2020self} & ResNet38 & I&31.9 \\
CONTA (NeurIPS20) \cite{zhang2020causal} & ResNet38 & I&32.8 \\
Kweon \textit{et al.} (ICCV21) \cite{kweon2021unlocking} & ResNet38 & I & 36.4 \\
CDA (ICCV21) \cite{su2021context} & ResNet38 & I &33.2 \\
PMM~(ICCV21) \cite{li2021pseudo}&ResNet38&I&36.7 \\
WS-FCN~(TNNLS23)~\cite{wang2023coupling} & ResNet38 & I & 27.5 \\
\midrule
SEC (CVPR16) \cite{kolesnikov2016seed} & VGG16 & I+S &22.4 \\
DSRG (CVPR18) \cite{huang2018weakly} & VGG16 & I+S&26.0 \\
DFN (TNNLS22)~\cite{zhang2022componentwise}&VGG16&I+S&26.8\\
EPS (CVPR21) \cite{lee2021railroad} & ResNet101 & I+S & 35.7\\
 \midrule
  \textbf{\nn} (ICCV21)~\cite{xu2021leveraging} & ResNet38 &I+S & 33.9\\
    \textbf{\nn+} (Ours) & ResNet38 &I+S & \textbf{37.0} \\
  \bottomrule 
\end{tabular}
\vspace{-1em}
\end{table}

\begin{table}[t]
\caption{Performance comparison of jointly learning different combinations of multiple tasks in terms of mIoU(\%) on PASCAL VOC \textit{val} set. $Seg., Cls.$, and $Sal.$ denote semantic segmentation, image classification, and saliency detection, respectively.}
\label{mt_results}
\vspace{-1em}
\centering
\small
\begin{tabular}{lcccc}
\toprule 
\multirow{2}{*}{Config}&\multicolumn{3}{c}{Branches}&\multirow{2}{*}{mIoU} \\\cline{2-4}
&Seg.&Cls.&Sal.&\\
\midrule
Baseline&\cmark&&&56.9 \\
\nn~(w/ seg., cls.)&\cmark&\cmark&&57.6 \\
\nn~(w/ seg., sal.)&\cmark&&\cmark&59.8\\
\nn~(w/ seg., cls., sal.)&\cmark&\cmark&\cmark&\textbf{60.8}\\

\bottomrule
\end{tabular}
\vspace{-1em}
\end{table}

\begin{figure}[t]
\setlength{\abovecaptionskip}{0cm}
\setlength{\belowcaptionskip}{0cm}
\begin{center}
\includegraphics[width=0.4\textwidth]{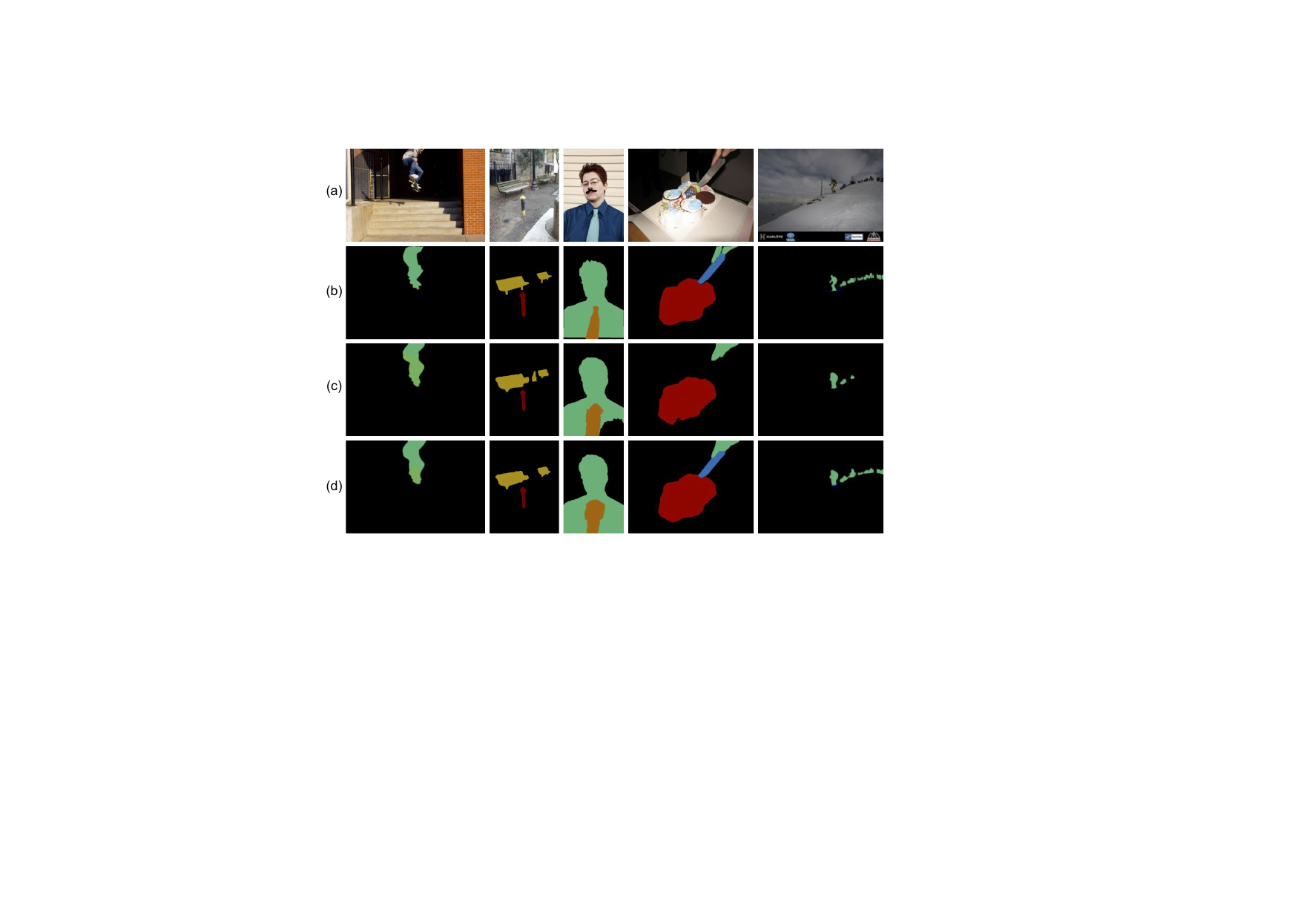}
\end{center}
 \vspace{-0.2cm}
  \caption{Qualitative segmentation results on the validation set of MS COCO. (a) Input images. (b) Ground-truth segmentation masks. (c) Predicted segmentation masks by~\nn. (d) Predicted segmentation masks by~\nn+.} 
\label{segresults}
\vspace{-0.2cm}
\end{figure}

\par\noindent\textbf{MS COCO.} To demonstrate the generalization ability of the proposed method, we also evaluated our method on the challenging MS COCO dataset. Note that a number of state-of-the-art methods~\cite{wu2021embedded,yao2021non} have only been tested on PASCAL VOC. Table \ref{coco} shows segmentation results of recent methods on the \textit{val} set.
The proposed \nn+ achieves a mIoU of 37.0\%, which is superior to the state-of-the-art methods. It outperforms \nn~and the recent method EPS~\cite{lee2021railroad} using off-the-shelf saliency maps by 3.1\% and 1.3\%, respectively. 
The proposed \nn+ achieves a reasonable performance gain compared to the recent WSSS method using only image-level labels. For example, Kweon~\textit{et al.}~\cite{kweon2021unlocking} proposed an adversarial erasing-based method to obtain improved CAM maps. However, it relies on an external method~\textit{i.e.,} PSA~\cite{ahn2018learning} to refine the improved CAM maps to generate high-quality pseudo labels.  PMM~\cite{li2021pseudo} proposed four strategies for the CAM map smoothing, pseudo label generation, noise-suppressing segmentation loss, and cyclic training, respectively. In contrast, our method proposes a new cross-task affinity learning method that can be used to refine CAM maps and enhance feature learning.  
However, (\textbf{i}) the CAM maps that the proposed affinities aim to refine are coarse and incomplete as they are extracted from an ordinary classifier, which makes the refinement more challenging compared to refining the improved CAM maps;
(\textbf{ii}) 
PMM~\cite{li2021pseudo} achieved an inferior result despite of using a segmentation loss function specifically designed for noisy pseudo labels. In contrast, our method used the standard segmentation loss function for the fully supervised semantic segmentation as in~\cite{jiangintegral, sun2020mining}. This treats noisy pseudo labels as ground truth and allows training a network to fit potential errors, thus possibly affecting the performance.
Fig.~\ref{segresults} presents several examples of qualitative segmentation results. The results of the proposed \nn+ (Fig.~\ref{segresults}d) are shown to be more precise than that of \nn~(Fig.~\ref{segresults}c) for different complex scenes, such as small objects or multiple instances.

\vspace{-1em}

\subsection{Ablation Analysis}
\label{sec:4.3}

\par\noindent\textbf{Effect of auxiliary tasks.} We compared results from the one-branch baseline model for semantic segmentation to several different variants: (\textbf{i}) baseline + cls: leveraging multi-label image classification, (\textbf{ii}) baseline + sal: leveraging saliency detection, and (\textbf{iii}) baseline + cls + sal: leveraging both image classification and saliency detection. Several conclusions can be drawn from Table \ref{mt_results}. Firstly, the baseline performance with the single task of semantic segmentation is only 56.9\%. Joint learning an auxiliary task of either image classification or saliency detection both improve the segmentation performance significantly. In particular, learning saliency detection brings a larger performance gain of around 3\%. Furthermore, leveraging both auxiliary tasks yields the best mIoU of 60.8\% without using any post-processing. This indicates that these two related auxiliary tasks can improve the representational ability of the network to achieve more accurate segmentation predictions in the weakly supervised scenario.

\begin{figure}[t]
\setlength{\abovecaptionskip}{0cm}
\setlength{\belowcaptionskip}{0cm}
\begin{center}
\includegraphics[width=0.4\textwidth]{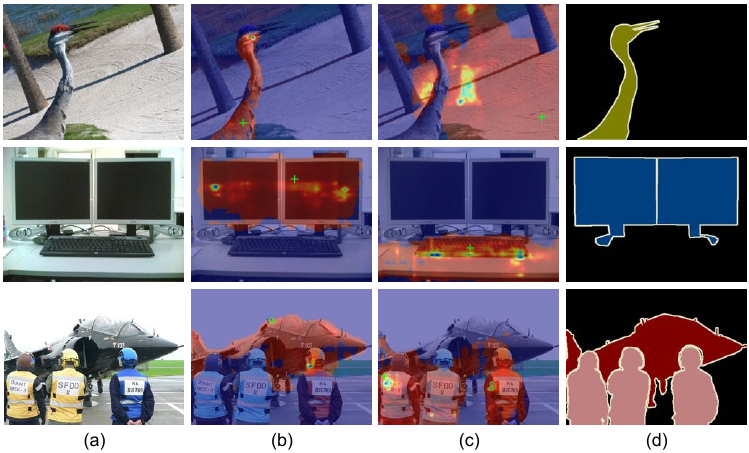}
\end{center}
 \vspace{-0.3cm}
  \caption{Visualization of the learned cross-task pairwise affinity maps of two selected points in the images on the PASCAL VOC \textit{train} set, by the proposed \nn. (a) Inputs. (b)-(c) Two learned cross-task affinity maps for two points marked by the green crosses. (d) Segmentation ground-truth.} 
\label{affmap}
\vspace{-1em}
\end{figure}

\begin{table}[t]
\caption{Effects of different pairwise affinity learning settings in terms of mIoU(\%) on PASCAL VOC \textit{val} set. CT: cross-task.}
\label{aff_results}
\vspace{-1em}
\centering
\small
\begin{tabular}{lc}
\toprule 
{Config}&mIoU \\
\midrule
\nn~(multi-task baseline)&60.8 \\
+ Seg. affinity with seg. constraint&61.5\\
+ CT affinity with seg. constraint&62.6\\
+ CT affinity with seg. and sal. constraints &\textbf{64.1}\\
\bottomrule
\end{tabular}
\vspace{-1em}
\end{table}

\vspace{0.5em}
\par\noindent\textbf{Different settings for affinity learning.} Table \ref{aff_results} shows ablation studies on the impact of different affinity learning settings on the segmentation performance. Without affinity learning, the segmentation mIoU is only 60.8\%. The performance is improved to 61.5\% by only learning segmentation affinity to refine segmentation predictions. Learning a cross-task affinity map that integrates both segmentation and saliency affinities brings a further performance boost of 1.1\%. By forcing the cross-task affinity map to learn to refine both segmentation and saliency predictions, our model attains a significant improvement, reaching a mIoU of 64.1\%. This shows the positive effect of the multi-task constraints on learning pixel affinities to enhance weakly supervised segmentation performance.

\begin{figure}[t]
\setlength{\abovecaptionskip}{0cm}
\setlength{\belowcaptionskip}{0cm}
\begin{center}
\includegraphics[width=0.35\textwidth]{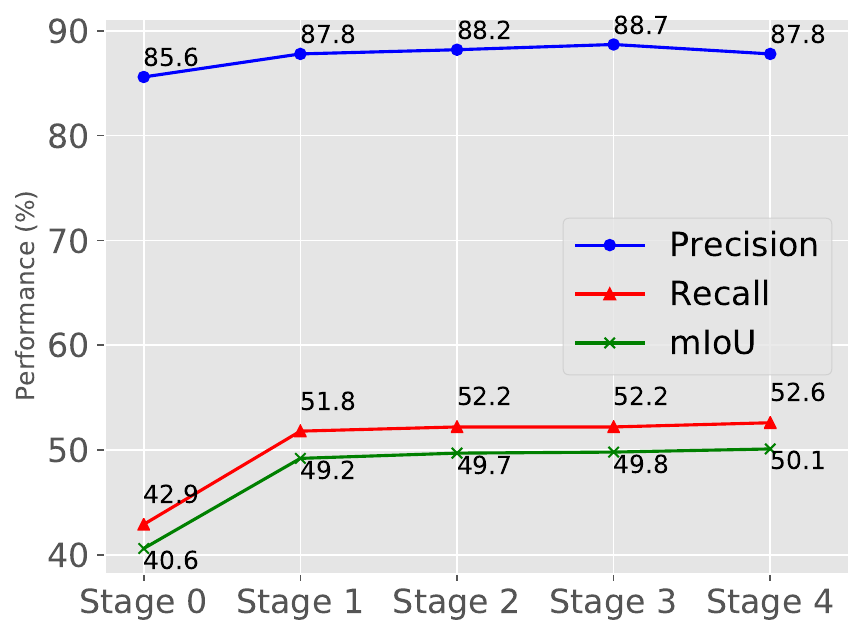}
\end{center}
 \vspace{-0.35cm}
  \caption{Evaluation of the segmentation PGT for each training stage in terms of precision(\%), recall(\%), and mIoU(\%) on PASCAL VOC \textit{train} set.} 
\label{pgt_line}
\vspace{-1.5em}
\end{figure}

Fig.~\ref{affmap} presents several examples of the learned cross-task affinity maps for two selected points in each image. The affinity map for each pixel in an image is of the image size and it represents the similarity of this pixel to all pixels in the image. We can observe that, in the first row, the affinity map of the point labeled as ``bird" highlights almost the entire object region although this point is far from the most discriminative object part ``head".
Moreover, in the second row, the learned affinity map for the point belonging to the monitor activates most regions of the two monitor instances while it does not respond to the ``keyboard" region which is similar in color in the background.
In the third row, the affinity maps for the ``airplane" and "person" points both present good boundaries.

\begin{figure*}[t]
\begin{center}
\includegraphics[width=0.83\textwidth]{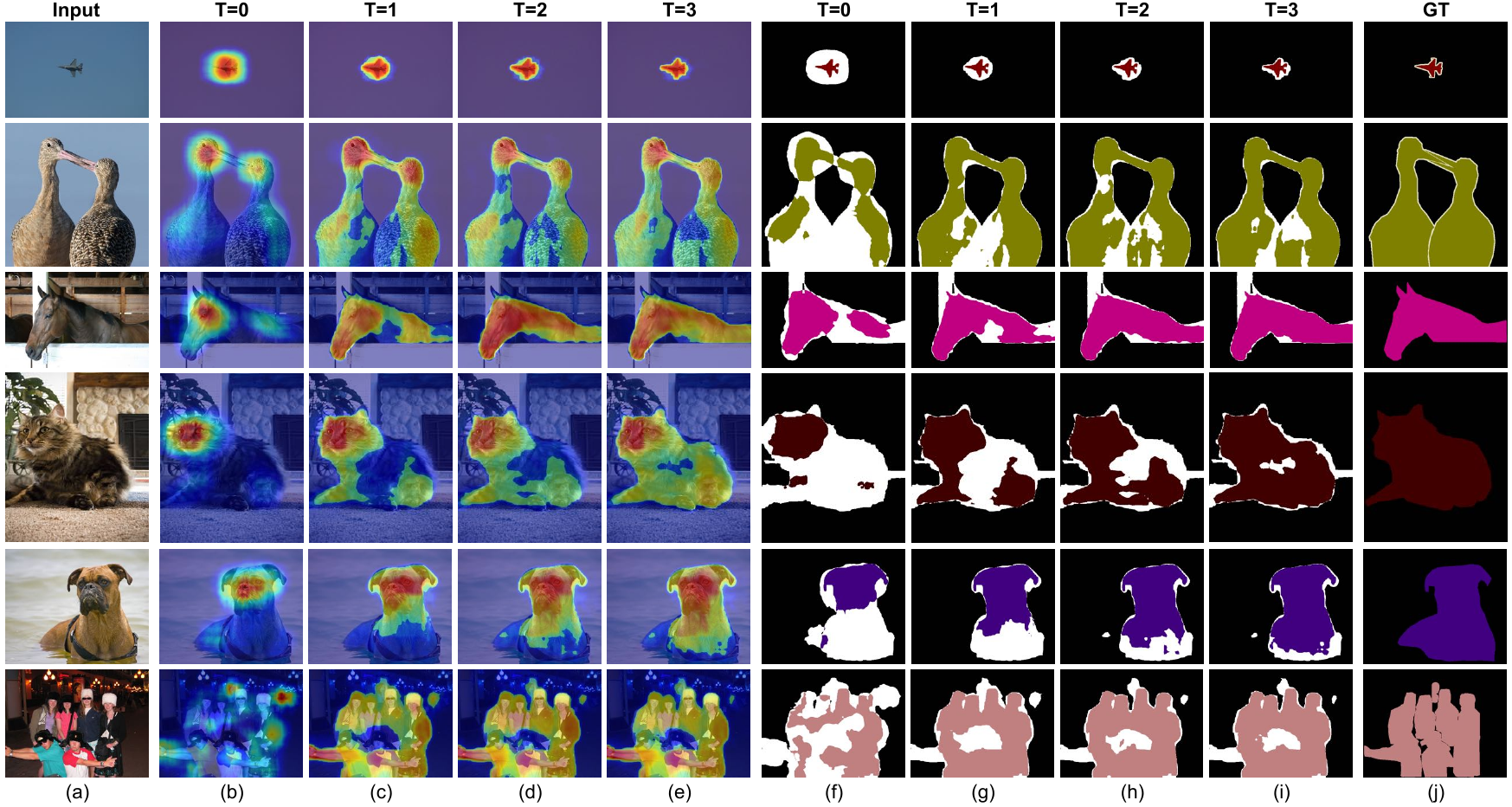}
\end{center}
\vspace{-1.5em}
  \caption{Visualization of CAM maps and segmentation PGT with iterative improvements. (a) Inputs. (b) Initial CAM maps without refinement. (c)-(e) Refined CAM maps used to generate pseudo segmentation labels for Stage 1 to Stage 3. (f) Initial pseudo semantic labels for Stage 0. (g)-(i) Pseudo segmentation labels for Stage 1 to Stage 3. (j) Segmentation ground-truth. With the iterative cross-task affinity learning, the refined CAM maps become more complete with more accurate boundaries and the generated corresponding pseudo-segmentation labels are closer to the ground truth in terms of precision and recall.} 
\label{cam_pgt}
\vspace{-0.4cm}
\end{figure*}

\begin{table}[t]
\caption{Segmentation performance of the proposed \nn~after different training stages in terms of mIoU (\%) on PASCAL VOC \textit{val} set. Stage 0 denotes the training stage with the initial pseudo-segmentation ground truth without refinement.}
\label{iterative_results}
\vspace{-1em}
\centering
\small
\begin{tabular}{lccccc}
\toprule 
&Stage0&Stage1&Stage2&Stage3&Stage4\\
\midrule
mIoU&64.1&65.6&66.0&\textbf{66.1}&\textbf{66.1} \\
\bottomrule
\end{tabular}
\vspace{-1em}
\end{table}

\vspace{0.5em}

\par\noindent\textbf{Iterative improvements.} To validate the effectiveness of the proposed iterative cross-task affinity learning, we evaluated the quality of the generated pseudo-segmentation labels for each training stage. As shown in Fig.~\ref{pgt_line}, the precision, recall, and mIoU of the initial pseudo segmentation labels generated by the CAM maps without refinement are 85.6\%, 42.9\%, and 40.6\%, respectively. After the first round of affinity learning, the updated pseudo labels for Stage 1 are significantly improved by 2.2\%, 8.9\%, and 8.6\% on these three metrics by using the refined CAM maps. Another round of affinity learning further boosts pseudo labels to 88.2\%, 52.2\%, and 49.7\% on the three metrics. In the following training stages, pseudo segmentation labels are slightly improved, and they are saturated at Stage 3 as we can observe that the precision starts to drop at Stage 4. As shown in Table \ref{iterative_results}, the segmentation performance presents consistent improvements as pseudo labels update after each training stage. Overall, the proposed \nn~attains a reasonable gain of 2\% by using iterative label updates with the learned cross-task affinity, reaching the best mIoU of 66.1\% without any post-processing.

To qualitatively evaluate the benefits of the proposed iterative affinity learning, Fig.~\ref{cam_pgt} presents several examples of CAM maps and the corresponding generated pseudo segmentation labels and their iterative improvements along the training stages. As shown in Fig.~\ref{cam_pgt} (b), the CAM maps without affinity refinement for the initial training stage are either over-activated for small-scale objects, sparse for multiple instances, or only focus on the local discriminative object parts for large-scale objects. By refining CAM maps with the cross-task affinity learned at Stage 0, Fig.~\ref{cam_pgt} (c) shows that more object regions are activated for large-scale objects and the CAM maps for small-scale objects are more aligned to object boundaries. With more training stages, as shown in Fig.~\ref{cam_pgt} (d)-(e), the refined CAM maps become more integral with more accurate boundaries, which is attributed to the more reliable affinity learned with iteratively updated PGT. The generated pseudo-segmentation labels are shown to be progressively improved in Fig.~\ref{cam_pgt} (f)-(i), and they are close to the ground-truth labels.

\begin{table}[t]
\caption{Segmentation results in terms of mIoU (\%) using saliency models pre-trained in supervised or weakly supervised forms on PASCAL VOC \textit{val} set. $^{*}$ denotes `without post-processing'.}
\vspace{-1em}
\resizebox{1.0\linewidth}{!}{
\begin{tabular}{lccccc}
\toprule
Pre-trained saliency models & Baseline & Final$^*$ & $\Delta$ & Final \\ 
\midrule
Zhang~\textit{et al.} (CVPR20)~\cite{zhang2020weakly}        &       53.0   &   63.7    &   10.7    & 66.5      \\
PoolNet (CVPR19)~\cite{liu2019simple}   &          55.7        &     65.7  &10.0 &68.4   \\
MINet (CVPR20)~\cite{pang2020multi}   &             55.8     &     66.6&10.8 & 68.9\\
Ours with DSS~\cite{hou2019deeply}          &       56.9   &  66.1     &  9.2     &     69.0 \\
\bottomrule
\end{tabular}
}
\label{fullsal}
\vspace{-1em}
\end{table}

\vspace{0.5em}
\par\noindent\textbf{Different pre-trained saliency models.} To evaluate the sensitivity to the pre-trained saliency models, we conducted experiments using different pre-trained saliency models, \textit{i.e.,} Zhang~\textit{et al.}~\cite{zhang2020weakly}, PoolNet~\cite{liu2019simple} used in~\cite{sun2020mining,zhang2020splitting}, and MINet~\cite{pang2020multi}, in which the first and the other two are in the weakly supervised and fully supervised forms, respectively. As shown in Table~\ref{fullsal}, our method achieves comparable results with these different saliency inputs, verifying the generalization ability of our method to different pre-trained saliency models. Moreover, with all these different pre-trained saliency models, our method can consistently produce significant performance improvements (see $\Delta$) over the baseline, which further confirmed the effectiveness of the proposed method.

\begin{figure}[t]
\begin{center}
\includegraphics[width=0.4\textwidth]{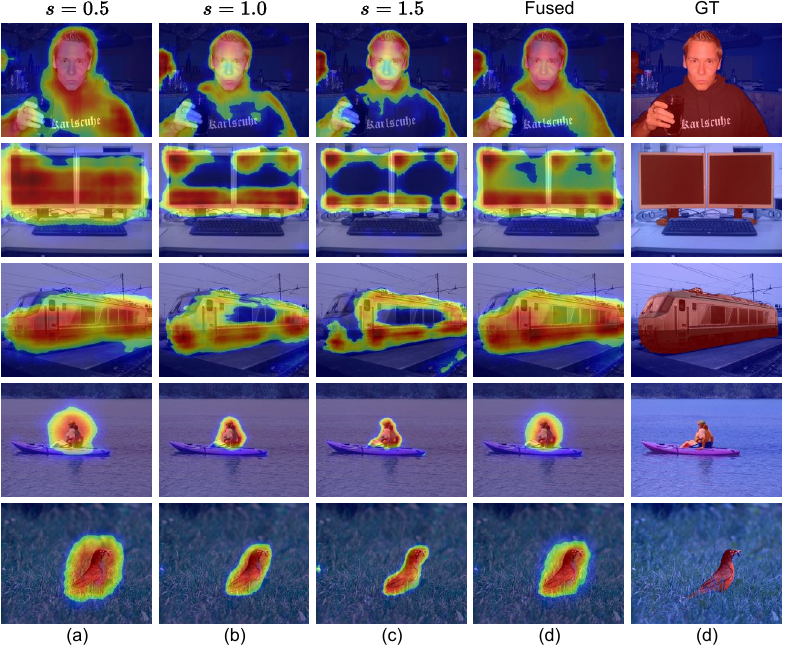}
\end{center}
 \vspace{-1em}
  \caption{Visualization of CAM maps at different scales (s denotes the scale).} 
\label{fig:mscam}
\vspace{-1em}
\end{figure}

\begin{table}[t]
\caption{Evaluation (\%) of the generated initial pseudo segmentation ground-truth labels using images at different scales on PASCAL VOC.}
\label{scales}
\vspace{-1em}
\centering
\small
\resizebox{\linewidth}{!}{
\begin{tabular}{lcccc}
\toprule 
Scale&Precision&Recall&mIoU& Time (approx.)\\
\midrule
(1.0)& 85.6&42.9&40.6& 12 mins\\
(1.0, 0.5, 1.5) &84.8 &\textbf{53.6}&\textbf{49.9}&35 mins\\
(1.0, 0.5, 1.5, 2.0) &\textbf{85.7}&50.8&47.7&71 mins\\

\bottomrule
\end{tabular}}
\vspace{-1em}
\end{table}

\begin{figure}[t]
\begin{center}
\includegraphics[width=0.4\textwidth]{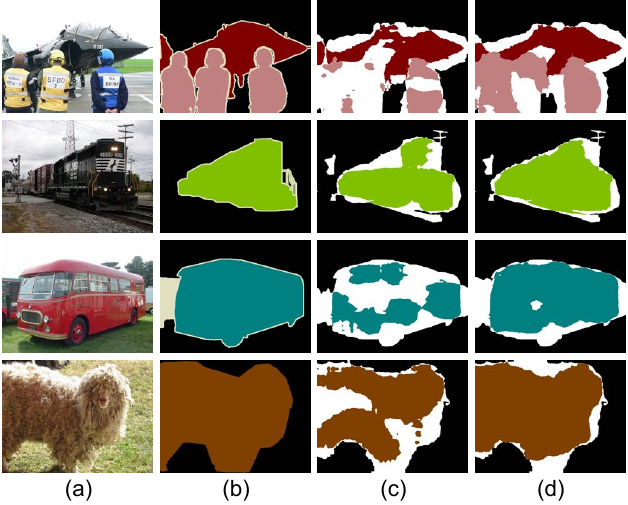}
\end{center}
 \vspace{-1em}
  \caption{Visualization of the initial pseudo segmentation ground-truth labels on the PASCAL VOC \textit{train} set (the white regions denote the unlabeled pixels, which do not contribute to the loss). (a) Inputs. (b) Ground-truth. (c) Pseudo-ground-truth labels using single-scale CAM maps. (d) Pseudo-ground-truth labels using multi-scale CAM maps.} 
\label{fig:mscam_pgt}
\vspace{-1.5em}
\end{figure}

\vspace{0.5em}

\par\noindent\textbf{Effect of multi-scale CAM.} 
As shown in Fig.~\ref{fig:mscam}, by using images at a smaller scale (\textit{e.g.}, 0.5), more complete CAM maps can be obtained. This is especially beneficial for medium- or large-sized objects, while it could also result in over-activated object regions for small-scale objects, bringing more false positives; In contrast, using images at a larger scale (\textit{e.g.}, 1.5) generally produces more accurate boundaries. This is particularly good for small objects, but it may lead to reduced object regions for large-scale objects. Therefore, using multi-scale inputs is useful to localize objects at different scales. Since CAM generally produces incomplete class-specific localization maps, using multi-scale inputs can significantly increase the recall of the pseudo-segmentation labels at a cost of a negligible drop in precision. 
Previous works~\cite{wang2020self,ahn2018learning,ahn2019weakly} commonly used inputs of scales (0.5, 1.0, 1.5, 2.0) to generate multi-scale CAM maps. However, in our case, as shown in Table~\ref{scales}, adding scale 2 only brings a minor gain of 0.9\% in precision but a drop of 2.8\% in recall and it is much more time-consuming, compared to only using the scales (0.5, 1.0, 1.5). Therefore, we only used the scales (0.5, 1.0, 1.5) in the experiments.
The visualization and the evaluation results of the generated pseudo segmentation labels by using single-scale and multi-scale CAM maps are shown in Fig.~\ref{fig:mscam_pgt} and Table~\ref{tab:mscam_pgt}, respectively. Fig.~\ref{fig:mscam_pgt} shows that using single-scale CAM maps to generate pseudo labels leads to a large proportion of unlabeled pixels especially for large-scale objects, such as the bus. Accordingly, the single-scale CAM-based pseudo labels have relatively low recall and mIoU scores of 42.9\% and 40.6\%, respectively. In contrast, by fusing the CAM maps generated from multiple images of different scales (\textit{i.e.,} scales of 0.5, 1.0, and 1.5), more object regions can be activated, contributing to more complete pseudo label maps (Fig.~\ref{fig:mscam_pgt}d) with significantly improved gains of 10.7\% and 9.3\% in terms of recall and mIoU, respectively.

\begin{table}[t]
\caption{Evaluation (\%) of the initial pseudo segmentation ground-truth labels when using single-scale and multi-scale CAM maps on PASCAL VOC.}
\label{tab:mscam_pgt}
\vspace{-1em}
\centering
\small
\begin{tabular}{lccc}
\toprule 
&Precision&Recall&mIoU \\
\midrule
Single-scale CAM&85.6&42.9&40.6 \\
Multi-scale CAM &84.8&53.6&49.9 \\
\bottomrule
\end{tabular}
\vspace{-1em}
\end{table}

\begin{figure}[t]
\begin{center}
\includegraphics[width=0.4\textwidth]{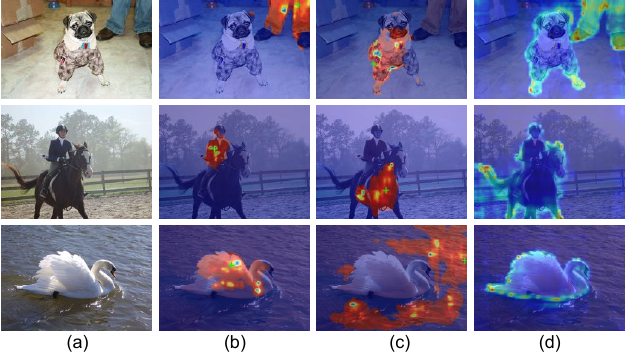}
\end{center}
 \vspace{-0.3cm}
  \caption{Visualization of the affinity maps by the proposed method on the PASCAL VOC \textit{train} set. (a) Inputs. (b)-(c) Pairwise affinity. (d) Unary affinity.} 
\label{fig:2affmap}
\vspace{-1em}
\end{figure}

\begin{table}[t]
\caption{Segmentation performance (\%) of the proposed~\nn~with different variants of the non-local block on the PASCAL VOC \textit{val} set.}
\label{nl_variants}
\vspace{-1em}
\centering
\small
\begin{tabular}{lc}
\toprule 
Config&mIoU \\
\midrule
\nn~w./ Pairwise affinity &64.1\\
\nn~w./ Disentangled Non-local~\cite{yin2020disentangled} &64.6\\
\nn+ w./ Dual affinity  & \textbf{65.2} \\
\bottomrule
\end{tabular}
\vspace{-1em}
\end{table}

\begin{table}[t]
\caption{Segmentation performance (\%) of the proposed \nn+ with different configurations.}
\label{segres_config}
\vspace{-1em}
\centering
\small
\resizebox{1.0\linewidth}{!}{
\begin{tabular}{lcc}
\toprule 
Config&Iterative Refinement&mIoU \\
\midrule
\nn~(Pairwise affinity+Single-scale CAM) &\xmark&64.1\\
\nn~(Pairwise affinity+Single-scale CAM) &\cmark&66.1\\
\midrule
\nn+ (Dual affinity+Single-scale CAM)&\xmark&65.2\\
\nn+ (Dual affinity+Multi-scale CAM)  &\xmark & 66.0 \\
\nn+ (Dual affinity+Multi-scale CAM)  &\cmark & \textbf{68.5} \\

\bottomrule
\end{tabular}}
\vspace{-1em}
\end{table}

\begin{table}[t]
\caption{Statistical results of paired-sample \emph{p}-test in terms of mIoU on the \textit{val} set of PASCAL VOC.}
\label{ptest}
\vspace{-1em}
\centering
\small
\resizebox{0.7\linewidth}{!}{
\begin{tabular}{lc}
\toprule 
AuxSegNet~\textit{v.s.} AuxSegNet+& \emph{p}-value \\
\midrule
w./o. iterative refinement & 5.25e-05 \\
w. iterative refinement & 1.04e-07 \\
\bottomrule
\end{tabular}}
\vspace{-1em}
\end{table}

\vspace{0.5em}
\par\noindent\textbf{Effect of the dual-affinity learning.} Table~\ref{nl_variants} compares the segmentation performance of the proposed~\nn~when using different affinity learning methods. The proposed \nn~with only the pairwise affinity learning enhances the features by aggregating query-dependent global context, achieving a mIoU of 64.1\%. In contrast, the proposed \nn+, which performs both pairwise and unary affinity learning to aggregate both query-dependent and query-independent global context, attains a performance gain of 1.1\%, achieving a mIoU of 65.2\%. Fig.~\ref{fig:2affmap} presents several visualization examples of the learned two types of affinity maps by the proposed cross-task dual-affinity learning module. As shown in Fig.~\ref{fig:2affmap} (b)-(c), the attention maps of different query positions (denoted by the green crosses) correctly highlight most regions of the corresponding object classes. 
In contrast, the query-independent attention maps (Fig.~\ref{fig:2affmap} d) learned from the unary affinity learning branch highlight accurate object boundaries.
In addition, we also compared the results between \nn+ and \nn~with the disentangled non-local block~\cite{yin2020disentangled}, which splits the original non-local block into a whitened pairwise term and a unary term. The proposed \nn+ outperforms the \nn~\textit{with} Disentangled Non-local by a margin of 0.6\%. We argue that the disentangled non-local block obtains pure pairwise attention (namely query-dependent attention) in the conventional learning framework without any supervisory constraints, by maximizing the difference between query and key features. In contrast, in the proposed framework, the learned pairwise affinity can capture the query-dependent relations by receiving supervision from pseudo labels of multiple dense prediction tasks during training. Therefore, the feature whitening process is redundant in the proposed learning framework and it may enlarge the difference of similar pixels hindering their affinity learning, which leads to a performance drop.

An overall comparison between \nn~and the proposed \nn+ is shown in Table~\ref{segres_config}. AuxSegNet proposed pairwise affinity learning and iterative refinement while only using single-scale CAM, achieving a mIoU of 66.1\%. In contrast, the proposed AuxSegNet+, which introduced dual affinity learning and multi-scale CAM, achieves a mIoU of 68.5\%, outperforming AuxSegNet by a margin of 2.4\%. The reason for this performance gain is two-fold: (\textbf{i}) the proposed dual affinity in AuxSegNet+ exploits both pairwise and unary affinities by aggregating both query-dependent and query-independent global context, which can further enhance the feature learning to achieve more accurate segmentation predictions, compared to AuxSegNet with only pairwise affinity learning.
(\textbf{ii}) the use of multi-scale CAM maps leads to more complete class-specific localization maps, generating higher-quality segmentation PGT, compared to using single-scale CAM. This thus provides better supervision to train the whole network and brings further performance gains. 
To demonstrate the significance of the improvements of the proposed AuxSegNet+ compared to AuxSegNet, we performed the paired-sample \emph{p}-test~\cite{wang2019thermal, bai2016pedestrian} on their segmentation results (mIoU). The paired-sample \emph{p}-test was conducted for both settings of not using and using iterative refinement. Table~\ref{ptest} shows that the proposed AuxSegNet+ outperforms AuxSegNet with a \emph{p}-value far below 0.05 (a commonly used significance threshold~\cite{chao2018learning}) in both settings. This indicates a statistically significant improvement in the proposed method.

\vspace{-1.5em}

\begin{figure}[t]
\begin{center}
\includegraphics[width=0.4\textwidth]{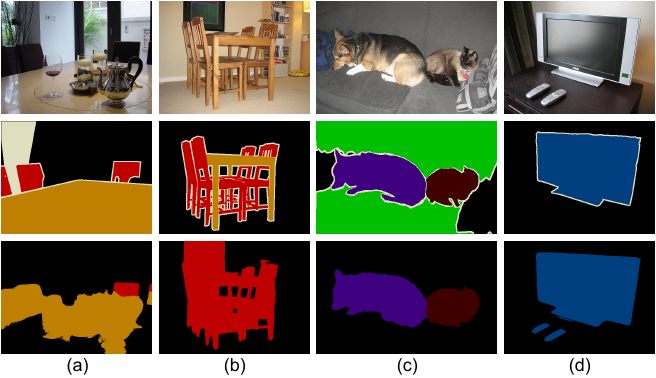}
\end{center}
 \vspace{-0.3cm}
  \caption{Failure cases of segmentation results on the PASCAL VOC \textit{val} set. Columns from top to bottom are input, GT masks and our results.} 
\label{fig:failurecases}
\vspace{-1.5em}
\end{figure}

\subsection{Discussion}
\label{sec:discussion}
\par\noindent\textbf{Failure analysis.}
Fig.~\ref{fig:failurecases} shows several representative failure cases of the segmentation results using the proposed method. In summary, the proposed method is prone to failures when several of the following challenging conditions co-exist in a scene: (\textbf{i}) Co-occurring objects or backgrounds; (\textbf{ii}) Objects without clear boundaries; (\textbf{iii}) Different semantic objects with similar appearance; (\textbf{iv}) Occluded objects. For instance, a desk and chairs often co-occur in a scene and they have similar appearances, such as color and legs, these two classes are thus prone to be misclassified into each other, especially when they are at proximate locations, as shown in Fig.~\ref{fig:failurecases} (a-b). Similarly, as shown in Fig.~\ref{fig:failurecases} (d), the silver TV remote controllers which belong to the background, were however misclassified as TV monitor, as there is a TV monitor with a silver frame in the image and they also often co-occur in the dataset. Besides, as shown in Fig.~\ref{fig:failurecases} (c), our method can hardly identify the sofa when it is not showing clear boundaries and is occluded by objects (\textit{i.e.,} the dog and the cat) sitting on it. 
\par\noindent\textbf{Future work.} Future research directions to overcome these challenges include: (\textbf{i}) One could exploit other sources, such as pre-trained vision-language (VL) models, to help generate high-quality pseudo-semantic segmentation labels. These VL models capture high-level semantic meanings of both image and language modalities. They are thus potentially helpful to reduce the contextual bias in the image classification or segmentation model due to the co-occurrences between objects or between objects and their backgrounds. 
(\textbf{ii}) More advanced neural network architectures, such as vision transformers (ViTs) can be used. ViTs can effectively model long-range dependencies, which are highly useful for tasks requiring understanding relationships between distant objects, such as semantic segmentation. 
Future works based on the proposed method include: (\textbf{i}) exploiting other relevant tasks that could further facilitate the recovery of accurate boundaries for both CAM maps and segmentation predictions, such as edge detection. (\textbf{ii}) extending the proposed method to video data.

\section{Conclusion}
\label{sec:conclusion}
In this work, we propose a weakly supervised multi-task framework, AuxSegNet+, which leverages auxiliary tasks of multi-label image classification and saliency detection to enhance the primary task of semantic segmentation with coarse and noisy supervisory sources. We propose a cross-task dual-affinity learning module, which fully exploits pixel affinities including pairwise and unary affinities from the saliency and segmentation feature maps. These two affinities can be used to enrich features and enhance predictions for saliency detection and semantic segmentation by aggregating the position-dependent and position-independent global contexts, respectively. The learned cross-task pairwise affinity map can be used to refine CAM maps for better pseudo-segmentation labels. Improved pseudo labels allow better training of the network to make more accurate segmentation predictions and learn more reliable pixel affinities. Iterative improvements in segmentation performance can be achieved by alternating the network training and label refinement processes. Extensive experimental results have demonstrated the effectiveness of the proposed method, 
which also achieved state-of-the-art results on the challenging PASCAL VOC and MS COCO.

This work proposes a novel WSSS framework that unifies CAM map generation and semantic segmentation prediction, enabling them to be mutually beneficial.
This provides a novel avenue for using multiple noisy sources to enrich features, enhance predictions, and refine pseudo labels for WSSS.
However, pairwise affinity learning in the proposed method is achieved by the non-local self-attention module, which has quadratic complexity. This could lead to expensive computations for large input resolutions. Future work could investigate more efficient methods for affinity learning.
The proposed method can generate high-quality pseudo-segmentation labels and produce accurate dense predictions using only limited and noisy supervision. This is highly useful in many applications, particularly in the fields of autonomous driving and robotics. Field managers in these areas could use the proposed method to improve their systems by automating accurate dense annotations of large datasets, which can save time and resources.


%





\ifCLASSOPTIONcaptionsoff
  \newpage
\fi



%


\bibliographystyle{IEEEtran}
\bibliography{newbib}

\begin{thebibliography}{10}
\providecommand{\url}[1]{#1}
\csname url@samestyle\endcsname
\providecommand{\newblock}{\relax}
\providecommand{\bibinfo}[2]{#2}
\providecommand{\BIBentrySTDinterwordspacing}{\spaceskip=0pt\relax}
\providecommand{\BIBentryALTinterwordstretchfactor}{4}
\providecommand{\BIBentryALTinterwordspacing}{\spaceskip=\fontdimen2\font plus
\BIBentryALTinterwordstretchfactor\fontdimen3\font minus \fontdimen4\font\relax}
\providecommand{\BIBforeignlanguage}[2]{{%
\expandafter\ifx\csname l@#1\endcsname\relax
\typeout{** WARNING: IEEEtran.bst: No hyphenation pattern has been}%
\typeout{** loaded for the language `#1'. Using the pattern for}%
\typeout{** the default language instead.}%
\else
\language=\csname l@#1\endcsname
\fi
#2}}
\providecommand{\BIBdecl}{\relax}
\BIBdecl

\bibitem{xu2021leveraging}
L.~Xu, W.~Ouyang, M.~Bennamoun, F.~Boussaid, F.~Sohel, and D.~Xu, ``Leveraging auxiliary tasks with affinity learning for weakly supervised semantic segmentation,'' in \emph{Int. Conf. Comput. Vis.}, 2021, pp. 6984--6993.

\bibitem{wang2020weakly}
G.~Wang, G.~Wang, X.~Zhang, J.~Lai, Z.~Yu, and L.~Lin, ``Weakly supervised person re-id: Differentiable graphical learning and a new benchmark,'' \emph{IEEE Trans. Neural Netw. Learn. Syst.}, vol.~32, no.~5, pp. 2142--2156, 2020.

\bibitem{zhang2020adapnet}
X.-Y. Zhang, C.~Li, H.~Shi, X.~Zhu, P.~Li, and J.~Dong, ``Adapnet: Adaptability decomposing encoder-decoder network for weakly supervised action recognition and localization,'' \emph{IEEE Trans. Neural Netw. Learn. Syst.}, early access, 23 Jan. 2020, doi:10.1109/TNNLS.2019.2962815.

\bibitem{yao2022ts}
Y.~Yao, F.~Wan, W.~Gao, X.~Pan, Z.~Peng, Q.~Tian, and Q.~Ye, ``Ts-cam: Token semantic coupled attention map for weakly supervised object localization,'' \emph{IEEE Trans. Neural Netw. Learn. Syst.}, pp. 1--13, early access, 23 Nov. 2022, doi:10.1109/TNNLS.2022.3218471.

\bibitem{zhang2022generalized}
D.~Zhang, G.~Guo, W.~Zeng, L.~Li, and J.~Han, ``Generalized weakly supervised object localization,'' \emph{IEEE Trans. Neural Netw. Learn. Syst.}, pp. 1--12, early access, 21 Sept. 2022, doi:10.1109/TNNLS.2022.3204337.

\bibitem{shen2018weakly}
Y.~Shen, R.~Ji, C.~Wang, X.~Li, and X.~Li, ``Weakly supervised object detection via object-specific pixel gradient,'' \emph{IEEE Trans. Neural Netw. Learn. Syst.}, vol.~29, no.~12, pp. 5960--5970, 2018.

\bibitem{zhang2020discriminant}
D.~Zhang, J.~Han, L.~Zhao, and T.~Zhao, ``From discriminant to complete: Reinforcement searching-agent learning for weakly supervised object detection,'' \emph{IEEE Trans. Neural Netw. Learn. Syst.}, vol.~31, no.~12, pp. 5549--5560, 2020.

\bibitem{wu2022enhanced}
Z.~Wu, J.~Wen, Y.~Xu, J.~Yang, X.~Li, and D.~Zhang, ``Enhanced spatial feature learning for weakly supervised object detection,'' \emph{IEEE Trans. Neural Netw. Learn. Syst.}, pp. 1--12, early access, 08 Jun. 2022, 10.1109/TNNLS.2022.3178180.

\bibitem{hu2018learning}
R.~Hu, P.~Doll{\'a}r, K.~He, T.~Darrell, and R.~Girshick, ``Learning to segment every thing,'' in \emph{IEEE Conf. Comput. Vis. Pattern Recog.}, 2018, pp. 4233--4241.

\bibitem{song2019box}
C.~Song, Y.~Huang, W.~Ouyang, and L.~Wang, ``Box-driven class-wise region masking and filling rate guided loss for weakly supervised semantic segmentation,'' in \emph{IEEE Conf. Comput. Vis. Pattern Recog.}, 2019, pp. 3136--3145.

\bibitem{lin2016scribblesup}
D.~Lin, J.~Dai, J.~Jia, K.~He, and J.~Sun, ``Scribblesup: Scribble-supervised convolutional networks for semantic segmentation,'' in \emph{IEEE Conf. Comput. Vis. Pattern Recog.}, 2016, pp. 3159--3167.

\bibitem{tang2018normalized}
M.~Tang, A.~Djelouah, F.~Perazzi, Y.~Boykov, and C.~Schroers, ``Normalized cut loss for weakly-supervised cnn segmentation,'' in \emph{IEEE Conf. Comput. Vis. Pattern Recog.}, 2018, pp. 1818--1827.

\bibitem{pathak2015constrained}
D.~Pathak, P.~Krahenbuhl, and T.~Darrell, ``Constrained convolutional neural networks for weakly supervised segmentation,'' in \emph{Int. Conf. Comput. Vis.}, 2015, pp. 1796--1804.

\bibitem{kolesnikov2016seed}
A.~Kolesnikov and C.~H. Lampert, ``Seed, expand and constrain: Three principles for weakly-supervised image segmentation,'' in \emph{Eur. Conf. Comput. Vis.}, 2016, pp. 695--711.

\bibitem{xu2021atrous}
L.~Xu, H.~Xue, M.~Bennamoun, F.~Boussaid, and F.~Sohel, ``Atrous convolutional feature network for weakly supervised semantic segmentation,'' \emph{Neurocomputing}, vol. 421, pp. 115--126, 2021.

\bibitem{zhang2022componentwise}
Z.~Zhang, Q.~Peng, S.~Fu, W.~Wang, Y.-M. Cheung, Y.~Zhao, S.~Yu, and X.~You, ``A componentwise approach to weakly supervised semantic segmentation using dual-feedback network,'' \emph{IEEE Trans. Neural Netw. Learn. Syst.}, pp. 1--14, early access, 04 Feb. 2022 2022, doi:10.1109/TNNLS.2022.3144194.

\bibitem{zhou2016learning}
B.~Zhou, A.~Khosla, A.~Lapedriza, A.~Oliva, and A.~Torralba, ``Learning deep features for discriminative localization,'' in \emph{IEEE Conf. Comput. Vis. Pattern Recog.}, 2016, pp. 2921--2929.

\bibitem{wei2017object}
Y.~Wei, J.~Feng, X.~Liang, M.-M. Cheng, Y.~Zhao, and S.~Yan, ``Object region mining with adversarial erasing: A simple classification to semantic segmentation approach,'' in \emph{IEEE Conf. Comput. Vis. Pattern Recog.}, 2017, pp. 1568--1576.

\bibitem{wei2018revisiting}
Y.~Wei, H.~Xiao, H.~Shi, Z.~Jie, J.~Feng, and T.~S. Huang, ``Revisiting dilated convolution: A simple approach for weakly-and semi-supervised semantic segmentation,'' in \emph{IEEE Conf. Comput. Vis. Pattern Recog.}, 2018, pp. 7268--7277.

\bibitem{jiangintegral}
P.-T. Jiang, Q.~Hou, Y.~Cao, M.-M. Cheng, Y.~Wei, and H.-K. Xiong, ``Integral object mining via online attention accumulation,'' in \emph{Int. Conf. Comput. Vis.}, 2019, pp. 2070--2079.

\bibitem{wang2020self}
Y.~Wang, J.~Zhang, M.~Kan, S.~Shan, and X.~Chen, ``Self-supervised equivariant attention mechanism for weakly supervised semantic segmentation,'' in \emph{IEEE Conf. Comput. Vis. Pattern Recog.}, 2020, pp. 12\,275--12\,284.

\bibitem{chaudhry2017discovering}
A.~Chaudhry, P.~K. Dokania, and P.~H. Torr, ``Discovering class-specific pixels for weakly-supervised semantic segmentation,'' in \emph{Brit. Mach. Vis. Conf.}, 2017, pp. 1--17.

\bibitem{hou2018self}
Q.~Hou, P.~Jiang, Y.~Wei, and M.-M. Cheng, ``Self-erasing network for integral object attention,'' in \emph{Adv. Neural Inform. Process. Syst.}, 2018, pp. 547--557.

\bibitem{sun2020mining}
G.~Sun, W.~Wang, J.~Dai, and L.~Van~Gool, ``Mining cross-image semantics for weakly supervised semantic segmentation,'' in \emph{Eur. Conf. Comput. Vis.}, 2020, pp. 347--365.

\bibitem{zhang2020splitting}
T.~Zhang, G.~Lin, W.~Liu, J.~Cai, and A.~Kot, ``Splitting vs. merging: Mining object regions with discrepancy and intersection loss for weakly supervised semantic segmentation,'' in \emph{Eur. Conf. Comput. Vis.}, 2020, pp. 663--679.

\bibitem{jiang2023metaseg}
S.~Jiang, J.~Li, Y.~Wang, W.~Wu, J.~Zhang, B.~Huang, and T.~Xu, ``Metaseg: Content-aware meta-net for omni-supervised semantic segmentation,'' \emph{{IEEE} Trans. Neural Networks Learn. Syst. (Early Access)}, pp. 1--13, 2023, doi: 10.1109/TNNLS.2023.3263335.

\bibitem{tang2018regularized}
M.~Tang, F.~Perazzi, A.~Djelouah, I.~Ben~Ayed, C.~Schroers, and Y.~Boykov, ``On regularized losses for weakly-supervised cnn segmentation,'' in \emph{Eur. Conf. Comput. Vis.}, 2018, pp. 507--522.

\bibitem{Keuniversal21}
T.~Ke, J.~Hwang, and S.~X. Yu, ``Universal weakly supervised segmentation by pixel-to-segment contrastive learning,'' in \emph{Int. Conf. Learn. Represent.}, 2021.

\bibitem{li2018tell}
K.~Li, Z.~Wu, K.-C. Peng, J.~Ernst, and Y.~Fu, ``Tell me where to look: Guided attention inference network,'' in \emph{IEEE Conf. Comput. Vis. Pattern Recog.}, 2018, pp. 9215--9223.

\bibitem{kweon2021unlocking}
H.~Kweon, S.-H. Yoon, H.~Kim, D.~Park, and K.-J. Yoon, ``Unlocking the potential of ordinary classifier: Class-specific adversarial erasing framework for weakly supervised semantic segmentation,'' in \emph{Int. Conf. Comput. Vis.}, 2021, pp. 6994--7003.

\bibitem{lee2021anti}
J.~Lee, E.~Kim, and S.~Yoon, ``Anti-adversarially manipulated attributions for weakly and semi-supervised semantic segmentation,'' in \emph{IEEE Conf. Comput. Vis. Pattern Recog.}, 2021, pp. 4071--4080.

\bibitem{zhang2020causal}
D.~Zhang, H.~Zhang, J.~Tang, X.~Hua, and Q.~Sun, ``Causal intervention for weakly-supervised semantic segmentation,'' in \emph{Adv. Neural Inform. Process. Syst.}, vol.~33, 2020, pp. 655--666.

\bibitem{su2021context}
Y.~Su, R.~Sun, G.~Lin, and Q.~Wu, ``Context decoupling augmentation for weakly supervised semantic segmentation,'' in \emph{Int. Conf. Comput. Vis.}, 2021, pp. 7004--7014.

\bibitem{xu2020scale}
L.~Xu, M.~Bennamoun, F.~Boussaid, and F.~Sohel, ``Scale-aware feature network for weakly supervised semantic segmentation,'' \emph{IEEE Access}, vol.~8, pp. 75\,957--75\,967, 2020.

\bibitem{yao2021non}
Y.~Yao, T.~Chen, G.-S. Xie, C.~Zhang, F.~Shen, Q.~Wu, Z.~Tang, and J.~Zhang, ``Non-salient region object mining for weakly supervised semantic segmentation,'' in \emph{IEEE Conf. Comput. Vis. Pattern Recog.}, 2021, pp. 2623--2632.

\bibitem{li2021group}
X.~Li, T.~Zhou, J.~Li, Y.~Zhou, and Z.~Zhang, ``Group-wise semantic mining for weakly supervised semantic segmentation,'' in \emph{{Proc. of AAAI Conference on Artificial Intelligence}}, vol.~35, no.~3, 2021, pp. 1984--1992.

\bibitem{wang2023coupling}
C.~Wang, D.~Zhang, L.~Zhang, and J.~Tang, ``Coupling global context and local contents for weakly-supervised semantic segmentation,'' \emph{IEEE Trans. Neural Netw. Learn. Syst.}, 2023.

\bibitem{chang2020weakly}
Y.-T. Chang, Q.~Wang, W.-C. Hung, R.~Piramuthu, Y.-H. Tsai, and M.-H. Yang, ``Weakly-supervised semantic segmentation via sub-category exploration,'' in \emph{IEEE Conf. Comput. Vis. Pattern Recog.}, 2020, pp. 8991--9000.

\bibitem{zhang2021complementary}
F.~Zhang, C.~Gu, C.~Zhang, and Y.~Dai, ``Complementary patch for weakly supervised semantic segmentation,'' in \emph{Int. Conf. Comput. Vis.}, 2021, pp. 7242--7251.

\bibitem{huang2018weakly}
Z.~Huang, X.~Wang, J.~Wang, W.~Liu, and J.~Wang, ``Weakly-supervised semantic segmentation network with deep seeded region growing,'' in \emph{IEEE Conf. Comput. Vis. Pattern Recog.}, 2018, pp. 7014--7023.

\bibitem{wang2018weakly}
X.~Wang, S.~You, X.~Li, and H.~Ma, ``Weakly-supervised semantic segmentation by iteratively mining common object features,'' in \emph{IEEE Conf. Comput. Vis. Pattern Recog.}, 2018, pp. 1354--1362.

\bibitem{ahn2018learning}
J.~Ahn and S.~Kwak, ``Learning pixel-level semantic affinity with image-level supervision for weakly supervised semantic segmentation,'' in \emph{IEEE Conf. Comput. Vis. Pattern Recog.}, 2018, pp. 4981--4990.

\bibitem{fan2020cian}
J.~Fan, Z.~Zhang, T.~Tan, C.~Song, and J.~Xiao, ``Cian: Cross-image affinity net for weakly supervised semantic segmentation,'' in \emph{{Proc. of AAAI Conference on Artificial Intelligence}}, vol.~34, no.~07, 2020, pp. 10\,762--10\,769.

\bibitem{xu2018pad}
D.~Xu, W.~Ouyang, X.~Wang, and N.~Sebe, ``Pad-net: Multi-tasks guided prediction-and-distillation network for simultaneous depth estimation and scene parsing,'' in \emph{IEEE Conf. Comput. Vis. Pattern Recog.}, 2018, pp. 675--684.

\bibitem{Sheng2019Unsupervised}
L.~Sheng, D.~Xu, W.~Ouyang, and X.~Wang, ``Unsupervised collaborative learning of keyframe detection and visual odometry towards monocular deep slam,'' in \emph{Int. Conf. Comput. Vis.}, 2019, pp. 4302--4311.

\bibitem{XuMoving}
D.~Xu, A.~Vedaldi, and J.~F.~Henriques, ``Moving slam: Fully unsupervised deep learning in non-rigid scenes,'' in \emph{{IEEE/RSJ Int. Conf. on Intelligent Robots and Systems (IROS)}}, 2021, pp. 4611--4617.

\bibitem{liu2019self}
S.~Liu, A.~Davison, and E.~Johns, ``Self-supervised generalisation with meta auxiliary learning,'' in \emph{Adv. Neural Inform. Process. Syst.}, 2019, pp. 1677--1687.

\bibitem{dai2016instance}
J.~Dai, K.~He, and J.~Sun, ``Instance-aware semantic segmentation via multi-task network cascades,'' in \emph{IEEE Conf. Comput. Vis. Pattern Recog.}, 2016, pp. 3150--3158.

\bibitem{chen2016dcan}
H.~Chen, X.~Qi, L.~Yu, and P.-A. Heng, ``Dcan: deep contour-aware networks for accurate gland segmentation,'' in \emph{IEEE Conf. Comput. Vis. Pattern Recog.}, 2016, pp. 2487--2496.

\bibitem{shen2019cyclic}
Y.~Shen, R.~Ji, Y.~Wang, Y.~Wu, and L.~Cao, ``Cyclic guidance for weakly supervised joint detection and segmentation,'' in \emph{IEEE Conf. Comput. Vis. Pattern Recog.}, 2019, pp. 697--707.

\bibitem{hwang2021weakly}
J.~Hwang, S.~Kim, J.~Son, and B.~Han, ``Weakly supervised instance segmentation by deep community learning,'' in \emph{IEEE Wint. Conf. App. Comput. Vis.}, 2021, pp. 1020--1029.

\bibitem{zhang2019reliability}
B.~Zhang, J.~Xiao, Y.~Wei, M.~Sun, and K.~Huang, ``Reliability does matter: An end-to-end weakly supervised semantic segmentation approach,'' in \emph{{Proc. of AAAI Conference on Artificial Intelligence}}, vol.~34, no.~07, 2020, pp. 12\,765--12\,772.

\bibitem{araslanov2020single}
N.~Araslanov and S.~Roth, ``Single-stage semantic segmentation from image labels,'' in \emph{IEEE Conf. Comput. Vis. Pattern Recog.}, 2020, pp. 4253--4262.

\bibitem{zeng2019joint}
Y.~Zeng, Y.~Zhuge, H.~Lu, and L.~Zhang, ``Joint learning of saliency detection and weakly supervised semantic segmentation,'' in \emph{Int. Conf. Comput. Vis.}, 2019, pp. 7223--7233.

\bibitem{lee2021railroad}
S.~Lee, M.~Lee, J.~Lee, and H.~Shim, ``Railroad is not a train: Saliency as pseudo-pixel supervision for weakly supervised semantic segmentation,'' in \emph{IEEE Conf. Comput. Vis. Pattern Recog.}, 2021, pp. 5495--5505.

\bibitem{hou2019deeply}
Q.~Hou, M.~Cheng, X.~Hu, A.~Borji, Z.~Tu, and P.~Torr, ``Deeply supervised salient object detection with short connections.'' \emph{IEEE Trans. Pattern Anal. Mach. Intell.}, vol.~41, no.~4, pp. 815--828, 2019.

\bibitem{lee2019ficklenet}
J.~Lee, E.~Kim, S.~Lee, J.~Lee, and S.~Yoon, ``Ficklenet: Weakly and semi-supervised segmentation using stochastic inference,'' in \emph{IEEE Conf. Comput. Vis. Pattern Recog.}, 2019, pp. 5267--5276.

\bibitem{vaswani2017attention}
A.~Vaswani, N.~Shazeer, N.~Parmar, J.~Uszkoreit, L.~Jones, A.~N. Gomez, {\L}.~Kaiser, and I.~Polosukhin, ``Attention is all you need,'' in \emph{Adv. Neural Inform. Process. Syst.}, 2017, pp. 5998--6008.

\bibitem{cao2019gcnet}
Y.~Cao, J.~Xu, S.~Lin, F.~Wei, and H.~Hu, ``Gcnet: Non-local networks meet squeeze-excitation networks and beyond,'' in \emph{Int. Conf. Comput. Vis. Worksh.}, 2019, pp. 1--10.

\bibitem{yin2020disentangled}
M.~Yin, Z.~Yao, Y.~Cao, X.~Li, Z.~Zhang, S.~Lin, and H.~Hu, ``Disentangled non-local neural networks,'' in \emph{Eur. Conf. Comput. Vis.}, 2020, pp. 191--207.

\bibitem{hu2018squeeze}
J.~Hu, L.~Shen, and G.~Sun, ``Squeeze-and-excitation networks,'' in \emph{IEEE Conf. Comput. Vis. Pattern Recog.}, 2018, pp. 7132--7141.

\bibitem{woo2018cbam}
S.~Woo, J.~Park, J.-Y. Lee, and I.~So~Kweon, ``Cbam: Convolutional block attention module,'' in \emph{Eur. Conf. Comput. Vis.}, 2018, pp. 3--19.

\bibitem{zhang2018occluded}
S.~Zhang, J.~Yang, and B.~Schiele, ``Occluded pedestrian detection through guided attention in cnns,'' in \emph{IEEE Conf. Comput. Vis. Pattern Recog.}, 2018, pp. 6995--7003.

\bibitem{zheng2017learning}
H.~Zheng, J.~Fu, T.~Mei, and J.~Luo, ``Learning multi-attention convolutional neural network for fine-grained image recognition,'' in \emph{Int. Conf. Comput. Vis.}, 2017, pp. 5209--5217.

\bibitem{everingham2010pascal}
M.~Everingham, L.~Van~Gool, C.~K. Williams, J.~Winn, and A.~Zisserman, ``The pascal visual object classes (voc) challenge,'' \emph{Int. J. Comput. Vis.}, vol.~88, no.~2, pp. 303--338, 2010.

\bibitem{lin2014microsoft}
T.-Y. Lin, M.~Maire, S.~Belongie, J.~Hays, P.~Perona, D.~Ramanan, P.~Doll{\'a}r, and C.~L. Zitnick, ``Microsoft coco: Common objects in context,'' in \emph{Eur. Conf. Comput. Vis.}, 2014, pp. 740--755.

\bibitem{chen2014semantic}
L.-C. Chen, G.~Papandreou, I.~Kokkinos, K.~Murphy, and A.~L. Yuille, ``Semantic image segmentation with deep convolutional nets and fully connected crfs,'' in \emph{Int. Conf. Learn. Represent.}, 2015.

\bibitem{hariharan2011semantic}
B.~Hariharan, P.~Arbel{\'a}ez, L.~Bourdev, S.~Maji, and J.~Malik, ``Semantic contours from inverse detectors,'' in \emph{Int. Conf. Comput. Vis.}, 2011, pp. 991--998.

\bibitem{luo2020learning}
W.~Luo and M.~Yang, ``Learning saliency-free model with generic features for weakly-supervised semantic segmentation.'' in \emph{{Proc. of AAAI Conference on Artificial Intelligence}}, vol.~34, no.~07, 2020, pp. 11\,717--11\,724.

\bibitem{sun2021ecs}
K.~Sun, H.~Shi, Z.~Zhang, and Y.~Huang, ``Ecs-net: Improving weakly supervised semantic segmentation by using connections between class activation maps,'' in \emph{Int. Conf. Comput. Vis.}, 2021, pp. 7283--7292.

\bibitem{li2021pseudo}
Y.~Li, Z.~Kuang, L.~Liu, Y.~Chen, and W.~Zhang, ``Pseudo-mask matters in weakly-supervised semantic segmentation,'' in \emph{Int. Conf. Comput. Vis.}, 2021, pp. 6964--6973.

\bibitem{qin2021activation}
J.~Qin, J.~Wu, X.~Xiao, L.~Li, and X.~Wang, ``Activation modulation and recalibration scheme for weakly supervised semantic segmentation,'' in \emph{{Proc. of AAAI Conference on Artificial Intelligence}}, vol.~36, no.~2, 2022, pp. 2117--2125.

\bibitem{fan2020learning}
J.~Fan, Z.~Zhang, C.~Song, and T.~Tan, ``Learning integral objects with intra-class discriminator for weakly-supervised semantic segmentation,'' in \emph{IEEE Conf. Comput. Vis. Pattern Recog.}, 2020, pp. 4283--4292.

\bibitem{wu2021embedded}
T.~Wu, J.~Huang, G.~Gao, X.~Wei, X.~Wei, X.~Luo, and C.~H. Liu, ``Embedded discriminative attention mechanism for weakly supervised semantic segmentation,'' in \emph{IEEE Conf. Comput. Vis. Pattern Recog.}, 2021, pp. 16\,765--16\,774.

\bibitem{wu2019wider}
Z.~Wu, C.~Shen, and A.~Van Den~Hengel, ``Wider or deeper: Revisiting the resnet model for visual recognition,'' \emph{Pattern Recognition}, vol.~90, pp. 119--133, 2019.

\bibitem{zhang2020weakly}
J.~Zhang, X.~Yu, A.~Li, P.~Song, B.~Liu, and Y.~Dai, ``Weakly-supervised salient object detection via scribble annotations,'' in \emph{IEEE Conf. Comput. Vis. Pattern Recog.}, 2020, pp. 12\,546--12\,555.

\bibitem{liu2019simple}
J.-J. Liu, Q.~Hou, M.-M. Cheng, J.~Feng, and J.~Jiang, ``A simple pooling-based design for real-time salient object detection,'' in \emph{IEEE Conf. Comput. Vis. Pattern Recog.}, 2019, pp. 3917--3926.

\bibitem{pang2020multi}
Y.~Pang, X.~Zhao, L.~Zhang, and H.~Lu, ``Multi-scale interactive network for salient object detection,'' in \emph{IEEE Conf. Comput. Vis. Pattern Recog.}, 2020, pp. 9413--9422.

\bibitem{ahn2019weakly}
J.~Ahn, S.~Cho, and S.~Kwak, ``Weakly supervised learning of instance segmentation with inter-pixel relations,'' in \emph{IEEE Conf. Comput. Vis. Pattern Recog.}, 2019, pp. 2209--2218.

\bibitem{wang2019thermal}
P.~Wang and X.~Bai, ``Thermal infrared pedestrian segmentation based on conditional gan,'' \emph{IEEE Trans. Image Process.}, vol.~28, no.~12, pp. 6007--6021, 2019.

\bibitem{bai2016pedestrian}
X.~Bai, P.~Wang, and F.~Zhou, ``Pedestrian segmentation in infrared images based on circular shortest path,'' \emph{IEEE Transactions on Intelligent Transportation Systems}, vol.~17, no.~8, pp. 2214--2222, 2016.

\bibitem{chao2018learning}
Y.-W. Chao, Y.~Liu, X.~Liu, H.~Zeng, and J.~Deng, ``Learning to detect human-object interactions,'' in \emph{IEEE Wint. Conf. App. Comput. Vis.}\hskip 1em plus 0.5em minus 0.4em\relax IEEE, 2018, pp. 381--389.

\end{thebibliography}

\vspace{-1em}
\begin{IEEEbiography}
[{\includegraphics[width=1in,height=1.25in,clip,keepaspectratio]{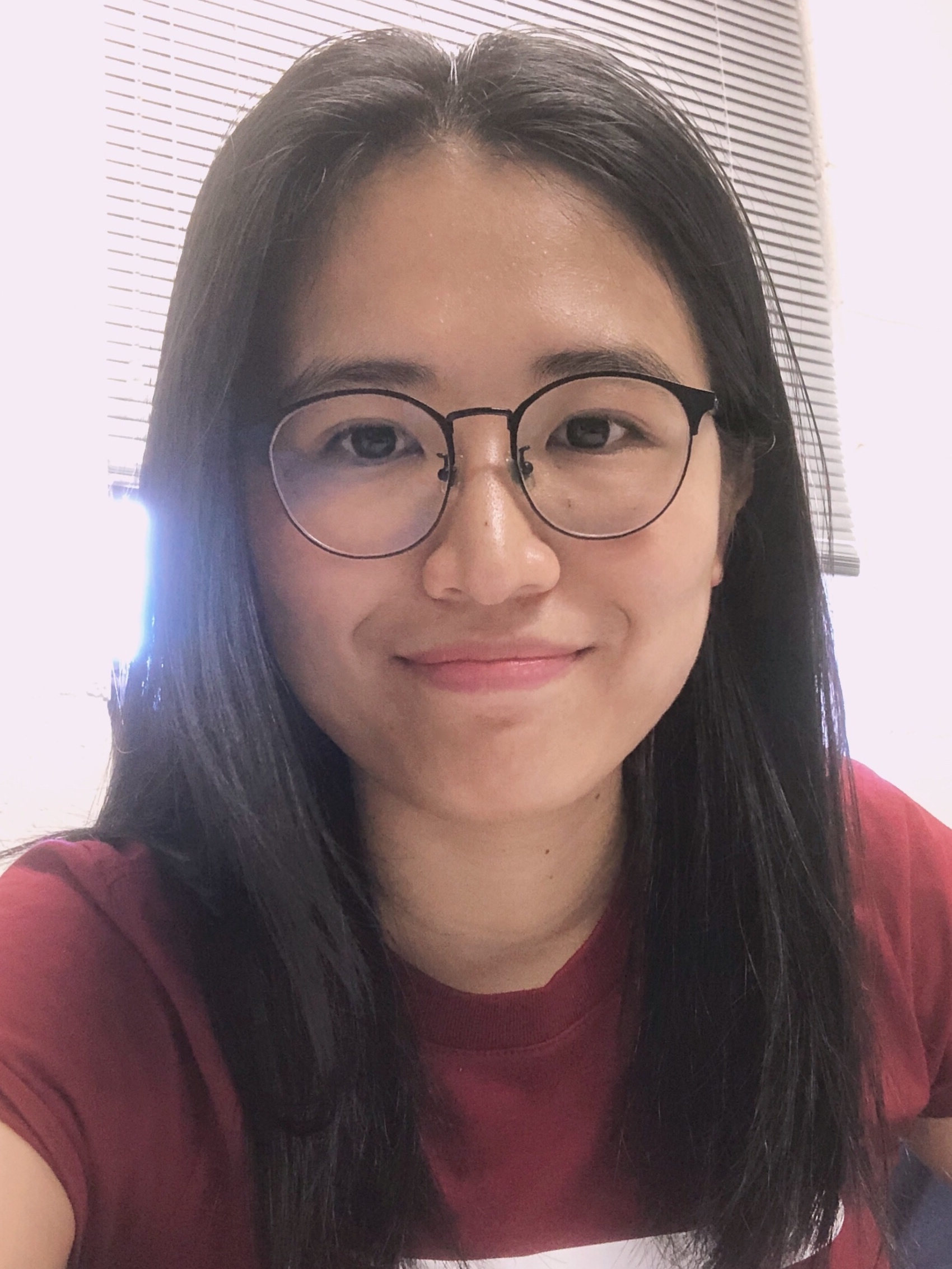}}]
{Lian Xu}
received a Ph.D. degree from the Department of Computer Science and Software Engineering at the University of Western Australia (UWA). She is currently a research fellow at the University of Western Australia. Her research interests include computer vision and machine learning.
\end{IEEEbiography}

\vspace{-1em}
\begin{IEEEbiography}[{\includegraphics[width=1in,height=1.25in,clip,keepaspectratio]{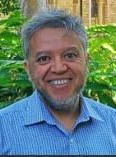}}]
{Mohammed Bennamoun} 
(Senior Member, IEEE) is a Winthrop Professor with the Department of Computer Science and Software Engineering at the University of Western Australia, Perth, Australia. He is a researcher in Computer Vision, Machine/Deep Learning, Robotics, and Signal/Speech Processing. He has co/authored four books, 14 book chapters, 200+ journal articles, and 270+ conference papers. He has won numerous research project grants from government and industry funding bodies.
\end{IEEEbiography}

\vspace{-1em}

\begin{IEEEbiography}[{\includegraphics[width=1in,height=1.25in,clip,keepaspectratio]{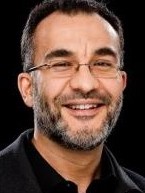}}]
{Farid Boussaid} received the M.S. and Ph.D. degrees in microelectronics from the National Institute of Applied Science (INSA), Toulouse, France, in 1996 and 1999 respectively. 
He joined the University of Western Australia, Crawley, Australia, in 2005, where he is currently a Professor. 
His current research interests include neuromorphic engineering, smart sensors, and machine learning.
\end{IEEEbiography}

\vspace{-1em}

\begin{IEEEbiography}[{\includegraphics[width=1in,height=1.25in,clip,keepaspectratio]{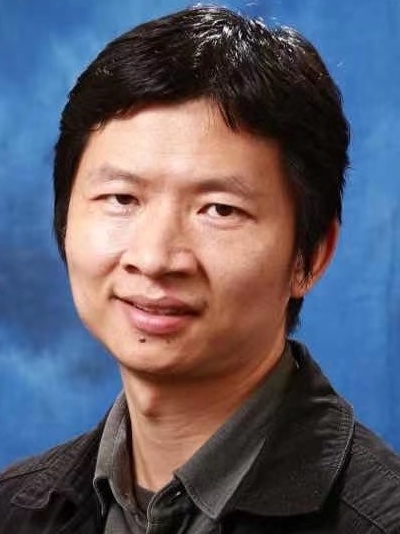}}]
{Wanli Ouyang} received
the Ph.D. degree from the Department of Electronic Engineering, The Chinese University of Hong Kong, Hong Kong, in 2010. He is currently a Professor with the Shanghai AI Laboratory, Shanghai, China. His research interests include image processing, computer vision, and pattern recognition.
\end{IEEEbiography}

\vspace{-1em}

\begin{IEEEbiography}[{\includegraphics[width=1in,height=1.25in,clip,keepaspectratio]{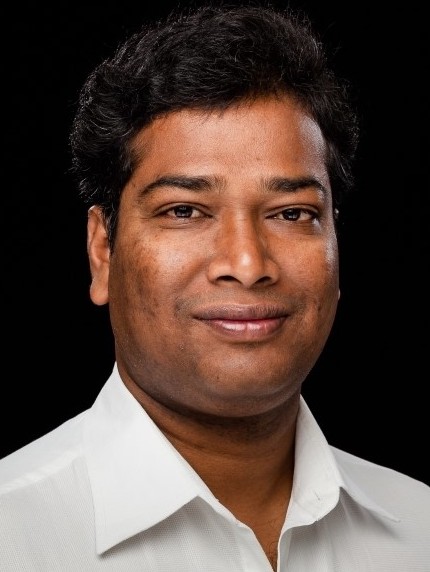}}]
{Ferdous Sohel} (M’08-SM’-14) received a PhD degree from Monash University, Australia. He is a Professor in Information Technology at Murdoch University, Australia. His research interests include computer vision, machine learning, digital agriculture and digital medicine \& health. 
He has published more than 220 scientific papers.
He is an Associate Editor of IEEE Transactions on Multimedia, IEEE Signal Processing Letters, and Computers and Electronics in Agriculture.

\end{IEEEbiography}

\vspace{-1.5em}
\begin{IEEEbiography}
[{\includegraphics[width=1in,height=1.25in,clip,keepaspectratio]{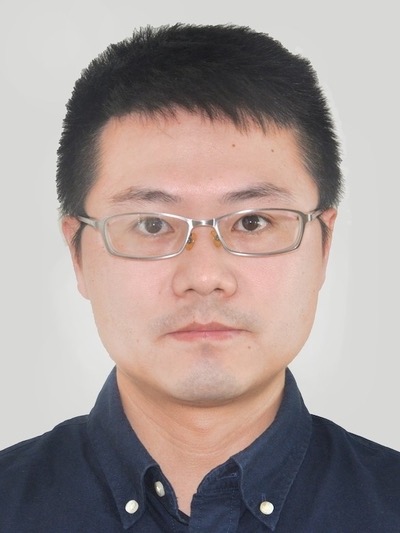}}]
{Dan Xu} received the Ph.D. degree from the University of Trento. He was a Postdoctoral Research Fellow in VGG at the University of Oxford. He is currently an Assistant Professor in the Department of Computer Science and Engineering at HKUST. He received the best scientific paper award at ICPR 2016, and a Best Paper Nominee at ACM MM 2018. He served as Area Chairs at multiple main-stream conferences including CVPR, AAAI, ACM Multimedia, WACV, ACCV and ICPR.
\end{IEEEbiography}

\end{document}